\documentclass{article}
\usepackage{amssymb}
\usepackage{graphicx}
\usepackage{amsmath}
\usepackage{booktabs}
\usepackage{hyperref}
\usepackage{multirow}
\usepackage{url}
\usepackage{makecell}
\usepackage{eso-pic}
\usepackage{xcolor}
\usepackage{cite}
\usepackage{booktabs} 
\usepackage{multirow} 
\usepackage{float}
\usepackage{booktabs}
\usepackage{multirow}
\usepackage{tabularx}
\usepackage{subcaption}
\usepackage{graphicx}
\usepackage{caption}
\usepackage{booktabs} 
\usepackage[utf8]{inputenc}
\usepackage{enumitem}
\usepackage{tcolorbox}
\tcbuselibrary{breakable}

\usepackage[preprint]{corl_2025} 
\usepackage{microtype}
\usepackage{etoolbox}
\newcommand{\MainTableSetup}{%
  \renewcommand{\arraystretch}{0.95}%
  \setlength{\tabcolsep}{4pt}%
}
\newcommand{\AppendixTableSetup}{%
  \renewcommand{\arraystretch}{1.0}%
  \setlength{\tabcolsep}{6pt}%
}
\MainTableSetup
\pretocmd{\appendix}{\AppendixTableSetup}{}{}

\usepackage{titlesec}
\titlespacing*{\section}{0pt}{0.8ex plus 0.2ex minus 0.2ex}{0.6ex plus 0.2ex minus 0.2ex}
\titlespacing*{\subsection}{0pt}{0.7ex plus 0.2ex minus 0.2ex}{0.5ex plus 0.2ex minus 0.2ex}
\titlespacing*{\subsubsection}{0pt}{0.6ex plus 0.2ex minus 0.2ex}{0.4ex plus 0.2ex minus 0.2ex}

\title{\fontsize{16}{18}\selectfont NavDreamer: Video Models as Zero-Shot 3D Navigators}

\author{
  \small
  Xijie Huang$^{\dagger, 1, 2}$  \And
  Weiqi Gai$^{\dagger, 2, 3}$ \And
  Tianyue Wu$^{1}$ \And
  Congyu Wang$^{1, 2}$ \And
  Zhiyang Liu$^{2}$ \And
  Xin Zhou$^{2}$ \AND
  Yuze Wu$^{\ast, 1, 2}$ \And
  Fei Gao$^{\ast, 1, 2}$ \AND
  Project page: \url{https://xinjiu612.github.io/NavDreamer}.
}

\begin{document}
\maketitle

\begingroup
\renewcommand\thefootnote{} 
\footnotetext{
  $^{\dagger}$ Equal contribution. \\ 
  $^{1}$ Zhejiang University, $^{2}$ Differential Robotics, $^{3}$ Beihang University. \\
  $^\ast$ Corresponding author: \texttt{wuyuze000@zju.edu.cn}, \texttt{fgaoaa@zju.edu.cn}. \\
}
\endgroup

\vspace{-8mm}
\begin{abstract}
Previous Vision-Language-Action models face critical limitations in navigation: scarce, diverse data from labor-intensive collection and static representations that fail to capture temporal dynamics and physical laws. We propose NavDreamer, a video-based framework for 3D navigation that leverages generative video models as a universal interface between language instructions and navigation trajectories. Our main hypothesis is that video's ability to encode spatiotemporal information and physical dynamics, combined with internet-scale availability, enables strong zero-shot generalization in navigation. To mitigate the stochasticity of generative predictions, we introduce a sampling-based optimization method that utilizes a VLM for trajectory scoring and selection. An inverse dynamics model is employed to decode executable waypoints from generated video plans for navigation. To systematically evaluate this paradigm in several video model backbones, we introduce a comprehensive benchmark covering object navigation, precise navigation, spatial grounding, language control, and scene reasoning. Extensive experiments demonstrate robust generalization across novel objects and unseen environments, with ablation studies revealing that navigation's high-level decision-making nature makes it particularly suited for video-based planning. 

\end{abstract}



\section{Introduction}
\label{sec:introduction}

The pursuit of general-purpose agents capable of generalization across diverse tasks and environments in the physical world is a long-cherished vision of the robotics and AI community. Within this scope, open-world navigation serves as a foundational requirement, demanding semantic contextualization and reactive spatial reasoning to decode the implicit information embedded within the natural world. Promisingly, Visual Language Models (VLLMs) \cite{team2023gemini, bai2023qwenvl, openai2023gpt4} have emerged with human-like reasoning, scene perception, and semantic analysis capabilities. Based on these advances, some robotics researchers have sought to incorporate action as a novel modality, leading to the development of Vision-Language-Action (VLA) models \cite{rt1}. These models typically employ VLMs as backbone for the sake of inheriting the spatial and linguistic understandings in VLMs, which have demonstrated preliminary generalization capabilities in specific scenarios \cite{kim2024openvla, pi0}. However, despite these advancements, the robotic domain seems yet to witness its own "aha moment" like LLM \cite{zeroshotlearner}.

A primary reason for the absence of this emergence point is that data scaling laws \cite{scaling,hernandez2021scaling} remain under-explored in robotics. While general-purpose intelligence typically surfaces from internet-scale datasets, robotic data collection remains a persistent challenge \cite{pi0.5, gen0}. Currently, high-fidelity demonstrations, from manipulation to navigation, all rely on labor-intensive and costly methods like bimanual teleoperation \cite{zhao2023learning} or manual remote control \cite{uavflow}, which inherently hinder the scalability of data collection. Furthermore, environmental and object diversity become increasingly instrumental for emergent intelligence once a certain data volume is reached \cite{scalelawrobo}. However, existing datasets \cite{openx, fang2023rh20t,uavflow} are still largely centered on constrained environments. Despite the achievement that specialized interfaces such as UMI \cite{umi} make in the wild collection, it is still hard to collect the datasets with the open-world's diversity. While simulation \cite{trackvla} and domain randomization \cite{viral} offer scalable data generation, the sim-to-real gap \cite{sim2real, gsnav} often leads to performance degradation when policies are deployed in real-world environments.

\begin{figure}
  \centering
  \includegraphics[width=1.0\linewidth]{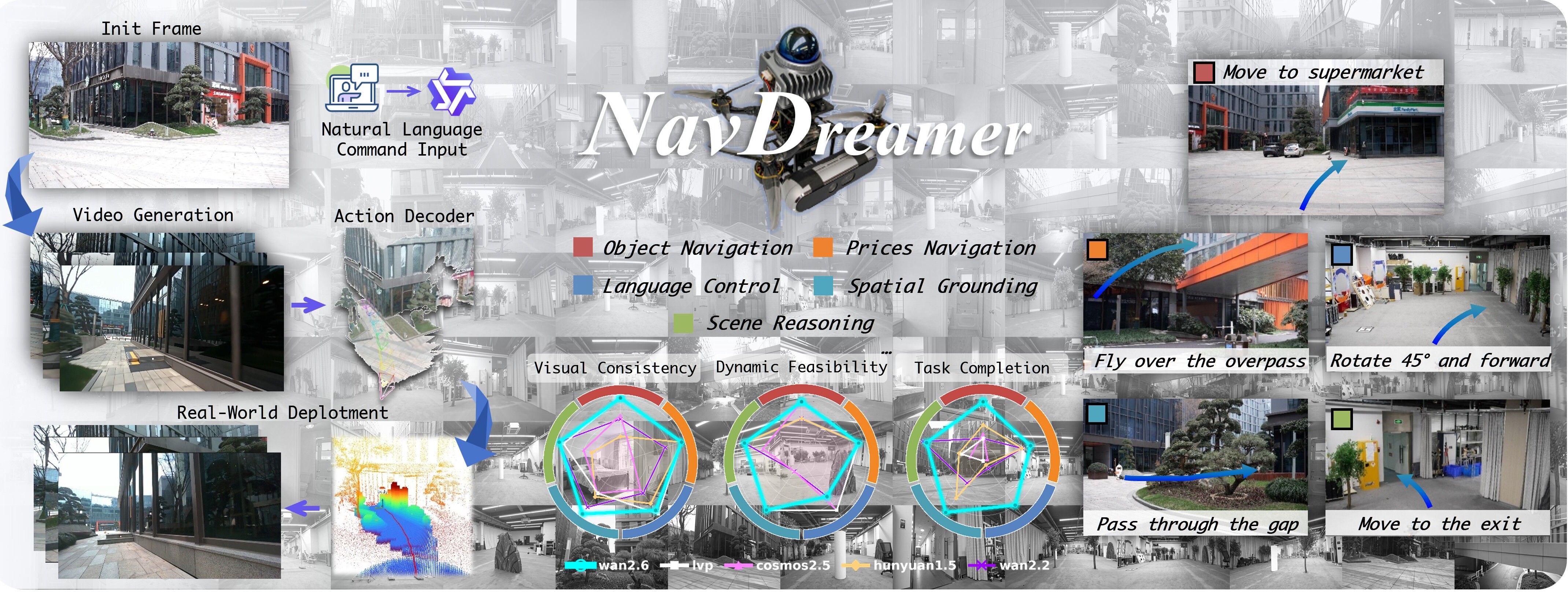}
    \caption{\textbf{System Overview.} \textbf{Left:} The pipeline leverages generative video models to transform text instructions and single RGB images into executable trajectories. \textbf{Right:} We evaluate the framework on five navigation task categories. }
  \label{fig:toutu}
  \vspace{-3mm}
\end{figure}


Beyond data scarcity, current VLA models are typically built upon vision-language backbones, which are trained on static image-text pairs \cite{cosmospolicy}. This static paradigm lacks the expressiveness to represent behaviors with strong temporal dependencies or those governed by physical laws, such as collisions and falling \cite{videolanguage}. Consequently, these models rely on a lossy representation that discards useful information during the transfer. This inherent information gap inevitably limits the model's capacity for reasoning world.


Recently, the rapid progress in generative video models \cite{wan, huanyuan, cosmos} has introduced a promising paradigm to address these challenges. First, unlike robotic action data, video datasets are available at an internet scale, enabling models to achieve significant zero-shot generalization across unseen tasks and environments \cite{goolevideo}. Second, video serves as a high-fidelity representation that captures intricate temporal and spatial information that language and static images often fail to resolve \cite{videolanguage}. These models have demonstrated an inherent understanding of physical dynamics \cite{goolevideo,videovla}, allowing them to reason about how the world evolves in response to specific movements. Furthermore, video functions as a unified state-action space, providing a universal interface to represent behaviors across diverse robotic platforms \cite{unipi}. This shared abstraction facilitates knowledge sharing by naturally capturing state-action trajectories within a consistent pixel space.



Driven by the superior generalization of video models, an increasing number of works have investigated transferring video priors to robotic policies \cite{cosmospolicy, dreamzero,mimic,lingbot}. Inspired by the paradigm of video-conditioned action decoding \cite{unipi,lvp}, we further explore their capabilities in the domain of 3D navigation. While previous manipulation-centric efforts were often hindered by the difficulty of decoding precise, low-level robotic actions \cite{andrychowicz2020learning,luo2025precise}, navigation primarily relies on high-level decision-making and directional guidance which is a more natural fit for video-based planning. But prior research in navigation has primarily focused on utilizing future frame prediction as an auxiliary signal to augment the navigation policy \cite{astranav, aerialworld}. In contrast, NavDreamer directly utilizes the generative video model as a high-level navigation planner. It serves as a unified interface that bridges the gap between high-level language instructions and executable navigation trajectories. To mitigate the stochasticity in generative models, we employ a sampling-based optimization strategy. Different from the energy-based (e.g., goal-image similarity) sampling methods \cite{navigationwm}, we use VLM to evaluate the video quality from different dimensions \cite{videohuman}. Then, we utilize the $\pi^3$ \cite{pi3} to decode executable waypoints from the generated video sequences. However, the physical scale estimation often degrades in outdoor environments. To address this, we incorporate metric depth priors from Moge2 \cite{moge2} to rectify the absolute scale ambiguity. The resulting calibrated waypoints are subsequently processed by a real-time reactive controller for flight execution. To evaluate the potential of video models in 3D navigation, NavDreamer demonstrates its powerful zero-shot generalization across diverse tasks involving novel objects and environments. Furthermore, to address the lack of video model benchmarks in 3D navigation, we introduce a comprehensive evaluation suite covering five dimensions: object navigation, precise navigation, spatial grounding, scene reasoning, and language control. Through extensive experiments and ablation studies, we provide key insights into the design choices for leveraging generative video models in 3D navigation.

In summary, our contributions are: (1) We propose zero-shot 3D navigators combining generative video model and inverse dynamics model to solve for high-level commands; (2) We design a comprehensive 3D navigation benchmark to evaluate state-of-the-art video models across five critical dimensions; (3) We explore various design choices through ablation studies to understand their effects on video-based 3D navigation performance.


\section{Related Work}
\label{sec:related work}

\subsection{Generative Video Models for Robotics}

The rapid advancement of generative video models \cite{cosmos,wan,huanyuan} has demonstrated remarkable zero-shot generalization to unseen scenarios \cite{goolevideo}, sparking a renewed interest in robotics. One line of research treats video models as high-fidelity neural simulation engines to synthesize large-scale, diverse robotic trajectories for policy pre-training \cite{dreamgen,gigabrain}. Another line utilizes video models as policy, which can be categorized into direct training and interface-based paradigms. In the context of direct training, some works integrate future frame prediction as a multimodal objective within the VLA paradigm to enforce physical consistency and foresight \cite{worldvla,dreamvla}. Others adopt pre-trained video models as VLA backbones to replace traditional VLMs, arguing that pre-trained video generators possess a more profound understanding of physical and spatiotemporal dynamics \cite{mimic, videovla,cosmos,dreamzero}. Alternatively, another paradigm treats video as a form of visual planning, where video models serve as an interface between visual imagination and robotic action. These methods first generate full-pixel future videos based on given conditions, subsequently extracting executable poses or trajectories from the generated frames \cite{lvp, unipi}. To extract continuous actions from these visual plans, several approaches have been proposed: one common method utilizes an Inverse Dynamics Model (IDM) to derive actions from frame sequences \cite{unipi}, while others map pixel-space motion directly into latent action spaces \cite{latentaction} via Vector-Quantized Variational Autoencoders (VQ-VAEs) \cite{VQ-VAEs}.

\subsection{3D Navigation and Autonomous Systems}

In recent years, UAV navigation has made significant strides across various specialized tasks, including drone racing \cite{naturedrone}, tracking \cite{track}, aerobatic \cite{teji}, and gap traversal \cite{zuankuang}. While these expert-based systems demonstrate high precision in specific domains, extending their capabilities to open-world semantic reasoning and broad task generalization remains a significant research objective. The emergence of VLA has introduced a unified paradigm to address these challenges. Notably, AerialVLN \cite{aerialvln} and CityNav \cite{citynav} established large-scale benchmarks for outdoor aerial navigation. FlightGPT \cite{flightgpt} further integrated reasoning-driven frameworks to enhance zero-shot generalization in complex scenarios. Frameworks like VLA-AN \cite{vla-an} utilize multi-stage training pipelines to improve scene understanding and navigation precision. However, these rely extensively on large human-annotated datasets for post-training VLM backbones. This reliance may pose challenges when transferring the generalization and semantic understanding of VLMs to the robotics domain, primarily due to the significant disparity in data scale. While training-free approaches \cite{spf} offer a more resource-efficient alternative, their application is currently often focused on relatively fundamental object-oriented navigation tasks, constrained by the rule-based strategies used to extract navigational cues from VLMs. \cite{aerialworld} leverages world models for future frame synthesis to provide similarity-based guidance for the policy, yet its deployment is restricted to constrained benchmarks such as image-goal navigation.


\section{Methods}
\label{sec:methods}

Given an input image $I$ and a language instruction $c$, NavDreamer first employs a video generation model to synthesize a predictive navigation sequence conditioned on these inputs. Notably, this generative process produces intuitive and information-dense visual cues that serve as high-level guidance. Qwen3-VL \cite{Qwen3VL} serves as a trajectory selector based on instruction alignment and environmental constraints like \cite{song2024vlm}. The selected video is subsequently downsampled at fixed intervals into an image sequence. We then employ $\pi^3$ \cite{pi3} to decode waypoints (e.g., x, y, z and yaw) from these images. To address the inherent scale ambiguity of $\pi^3$ in outdoor environments, we apply a robust scale recovery method using metric depth estimates as a reference. These calibrated waypoints serve as the foundation for the low-level planning module, where the final trajectory is optimized in accordance with the real-time obstacle distribution.

\begin{figure}[h]
  \centering
  \includegraphics[width=0.98\linewidth]{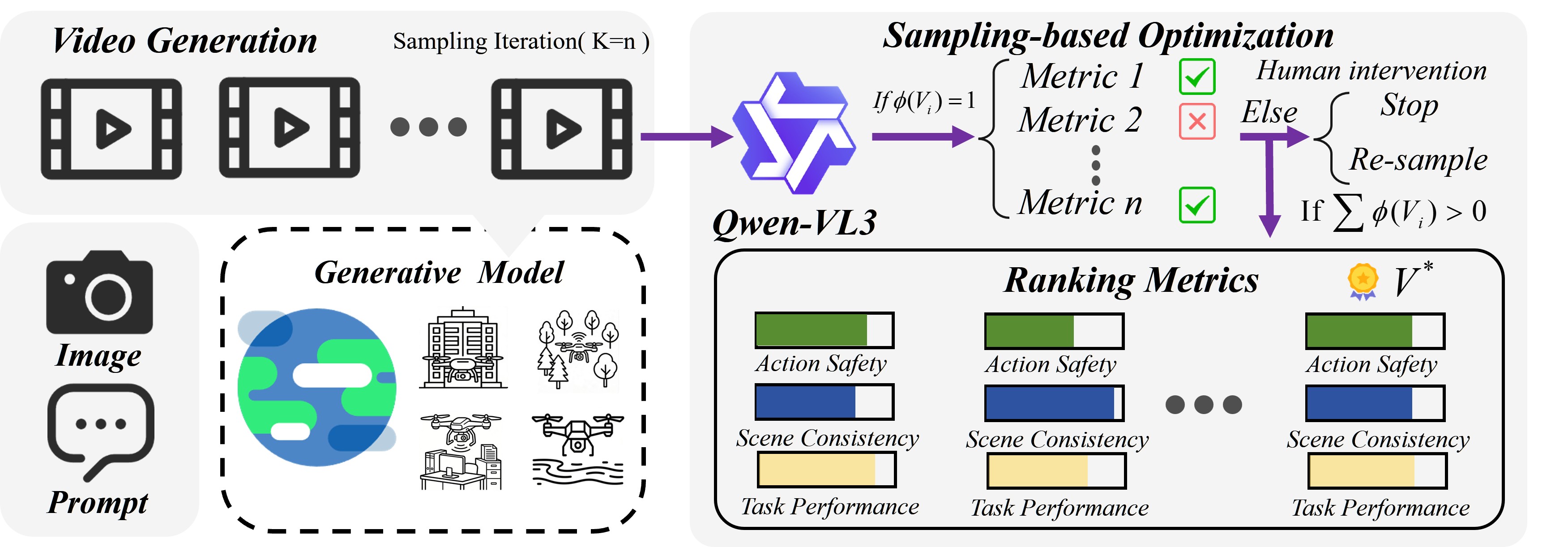}
    \caption{\textbf{Framework of Optimization through Generative Sampling.} The model generates $K$ video samples, which are evaluated by Qwen-VL3 on action safety, scene consistency, and task performance. If no valid samples exist ($\sum \phi(V_i) = 0$), human intervention determines whether to re-sample to obtain the optimal video $V^*$.}
  \label{fig:optimaztion}
  \vspace{-1mm}
\end{figure}

\subsection{Optimization through Generative Sampling}

Enhancing interpretability is a core objective in learning-based robotics. Given an observation $I_0$ and instruction $c$, the generative process is defined as $\mathcal{V} \sim P(\mathcal{V} \mid I_0, c)$. Since video generation is stochastic and identical conditions may yield divergent outcomes, we generate $K$ independent candidates $\{\mathcal{V}_1, \dots, \mathcal{V}_K\}$ to leverage sample diversity and mitigate instability. To ensure reliability, Qwen3-VL \cite{Qwen3VL} is utilized to evaluate these samples. As shown in Fig.~\ref{fig:optimaztion}, we define a binary success indicator $\Phi(\mathcal{V}_i) \in \{0, 1\}$ to verify whether a candidate sequence represents a successful and safe execution:

\begin{equation}
\Phi(\mathcal{V}_i) = 
\begin{cases} 
1, & \text{if } \text{VLM-Judge}(\mathcal{V}_i) \text{ is } \textit{True} \\
0, & \text{otherwise}
\end{cases}
\end{equation}

where a true result signifies that the video is both task-consistent and collision-free. This binary filter forms the set of valid candidates $\mathcal{V}_{\mathrm{valid}} = \{ \mathcal{V}_i \mid \Phi(\mathcal{V}_i) = 1 \}$. Then, we perform a fine-grained evaluation based on three metrics: action safety ($\text{sc}_{as}$), scene consistency ($\text{sc}_{sc}$), and task performance ($\text{sc}_{tp}$). The overall quality score $\mathcal{R}(\mathcal{V}_i)$ is calculated via a weighted sum:
\begin{equation}
\mathcal{R}(\mathcal{V}_i) = w_{\mathrm{as}} \cdot \mathrm{sc}_{\mathrm{as}} + w_{\mathrm{sc}} \cdot \mathrm{sc}_{\mathrm{sc}} + w_{\mathrm{tp}} \cdot \mathrm{sc}_{\mathrm{tp}}
\end{equation}

where $w_{as}, w_{sc}, w_{tp}$ represent the predefined importance weights for each dimension, set to 0.8, 0.8, 1.4, respectively. The final optimal video $\mathcal{V}^*$ is selected as the sample that maximizes the reward:
\begin{equation}
\mathcal{V}^* = \arg\max\nolimits_{\mathcal{V}_i \in \mathcal{V}_{\mathrm{valid}}} \mathcal{R}(\mathcal{V}_i)
\end{equation}

This mechanism effectively emulates human cognitive foresight, prioritizing the exclusion of "failed futures" before optimizing for execution quality. If $\mathcal{V}_{\mathrm{valid}}$ is null, the system issues a signal to a human supervisor to decide whether to resample or terminate the operation.

\subsection{High-Level Action From Video}



\begin{figure}[h]
  \centering
  \includegraphics[width=0.98\linewidth]{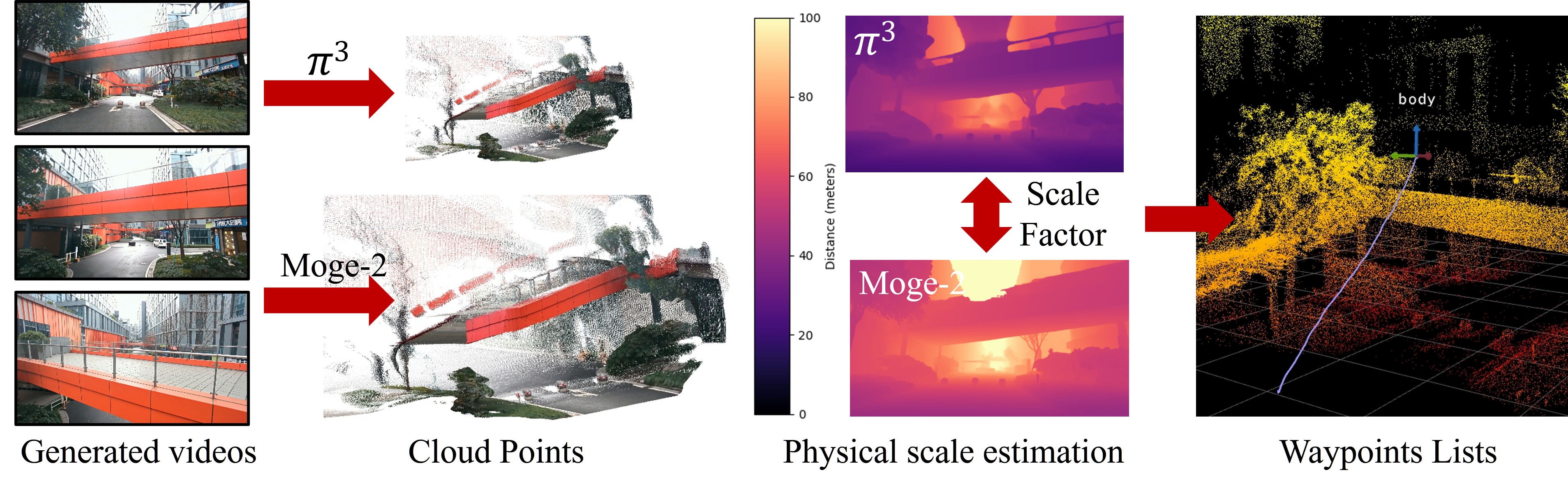}
    \caption{\textbf{High-level Waypoint Decoding and Metric Scale Alignment.} $\pi^3$ extracts initial waypoints and point clouds from generated RGB videos, while Moge-2 provides metric-scale depth references to resolve scale ambiguity. A global scale factor derived from cross-referencing both models transforms synthesized trajectories into physically grounded waypoint lists.}
  \label{fig:action_decoder}
\end{figure}

To bridge the gap between the normalized geometry predicted by $\pi^3$ \cite{pi3} and the metric physical world, we propose a robust scale recovery module. Given an image sequence $\{I_t\}_{t=1}^N$, we first employ a metric depth estimator (e.g., Moge2 \cite{moge2}) to obtain the reference depth maps $D^{ref}_t$ as shown in Fig.~\ref{fig:action_decoder}. Concurrently, $\pi^3$ is utilized to generate a sequence of local pointmaps $X_t \in \mathbb{R}^{H \times W \times 3}$. For each frame $t$, we define a valid observation mask $M_t$ to filter out outliers and extreme values (e.g., sky or extremely far objects), typically constrained within a reliable sensing range $\tau \in [0.5, 30]$ meters:


\begin{equation}
M_t = \{ (u, v) \mid \tau_{\mathrm{min}} < D^{\mathrm{ref}}_t(u,v) < \tau_{\mathrm{max}} \;\text{and}\; Z^{\mathrm{pred}}_t(u,v) > 0 \}    
\end{equation}
where $Z^{pred}_t(u, v)$ represents the depth values derived from the $\pi^3$-generated point maps. The per-pixel scale ratio $s_t(u,v)$ is computed within the valid region defined by $M_t$. To ensure robustness against depth noise and geometric misalignments, we adopt a median-based consensus to estimate the global scale factor $\mathcal{S}$:$$\mathcal{S} = \text{median} \left( \bigcup_{t=1}^N \left\{ \frac{D^{\mathrm{ref}}_t(u,v)}{Z^{\mathrm{pred}}_t(u,v)} \right\}_{(u,v) \in M_t} \right)$$Finally, the normalized waypoints $\mathbf{w}_t \in \mathbb{R}^3$ extracted from $\pi^3$ are transformed into the metric space as $\mathbf{W}_t = \mathcal{S} \cdot \mathbf{w}_t$, ensuring the decoded trajectory aligns with the physical ground truth for low-level navigation.

\subsection{Low-level Trajectory Generation}

Since the waypoints derived from video synthesis may contain errors in absolute physical scale, applying them directly to a position controller is risky. To ensure the system safely handles estimation inaccuracies and real-world uncertainties, we use Ego-Planner \cite{ego} as our low-level navigation module. This module receives a 3D position goal and plans a collision-free trajectory in real time. The system monitors the Euclidean distance to the current target; once this distance falls below a certain threshold, the next waypoint in the list is sent to the planner. This approach ensures that the total flight path combines the intended behaviors from the high-level video model with the safety of real-time obstacle avoidance. 

To ensure the physical flight execution reflects the implicit dynamic constraints captured in the synthesized video, we first calculate the maximum velocity and acceleration from the waypoint sequence. These values are then applied as adaptive constraints during the trajectory optimization process to match the video's dynamics. Additionally, we implement independent control for the yaw angle to make the system easier to operate. This real-time planning approach ensures that the final flight path is both robust and safe in complex environments.


\section{Experiments}
\label{sec:experiment}
To fully evaluate the potential of video models in 3D navigation, we conducted extensive experiments and ablation studies focusing on four questions: (1) What is the comparative efficacy of current state-of-the-art open-source versus closed-source video models in multifaceted navigation tasks? (2) How can precise actions be extracted from synthesized videos? (3) How does the system perform in real-world deployments and how well does it generalize? (4) In what ways do the sampling-based optimization method and prompt levels impact the reliability of the generated trajectories?

To achieve this, we first designed a suite of diverse tasks and suitable metrics for evaluating 3D navigation, because the most existing metrics for video models only focus on visual clarity and temporal smoothness. We then benchmarked the performance of several state-of-the-art open-source and closed-source models on these tasks. For action extraction, we performed ablation studies on the scale recovery components to demonstrate their ability to resolve the scale ambiguity. Through diverse tasks from indoor to outdoor, we prove that video models have zero-shot generation in unseen scenes. Finally, we analyzed the impact of sampling iterations on the success rate and examined how different prompt engineering strategies affect the stability and accuracy of the generated videos.

\vspace{1mm} 
\begin{table}[htbp]
  \centering
  \scriptsize 
  \setlength{\tabcolsep}{3pt} 
  \renewcommand{\arraystretch}{1.1} 
  
  \begin{tabular}{ll cccc c}
    \toprule
    & & \multicolumn{4}{c}{\textbf{Open-source}} & \textbf{Closed} \\
    \cmidrule(lr){3-6} \cmidrule(lr){7-7}
    \textbf{Task} & \textbf{Metric} & \textbf{Wan 2.2 \cite{wan}} & \textbf{Hunyuan 1.5 \cite{huanyuan}} & \textbf{Cosmos 2.5 \cite{cosmos}} & \textbf{LVP \cite{lvp}} & \textbf{Wan 2.6} \\
    \midrule
    
    \multirow{3}{*}{\shortstack[l]{\textbf{Object}\\\textbf{Navigation}}} 
      & Visual Consistency & 0.67 & 0.40 & 0.73 & \textbf{1.00} & \textbf{1.00} \\
      & Dynamic Feasibility & 0.93 & 0.93 & \textbf{1.00} & \textbf{1.00} & \textbf{1.00} \\
      & Task Completion & 0.33 & 0.53 & 0.40 & 0.33 & \textbf{0.87} \\
    \midrule
    
    \multirow{3}{*}{\shortstack[l]{\textbf{Precise}\\\textbf{Navigation}}} 
      & Visual Consistency & 0.73 & 1.00 & 0.20 & \textbf{1.00} & \textbf{1.00} \\
      & Dynamic Feasibility & \textbf{1.00} & \textbf{1.00} & \textbf{1.00} & \textbf{1.00} & \textbf{1.00} \\
      & Task Completion & 0.60 & 0.60 & 0.07 & 0.00 & \textbf{0.73} \\
    \midrule

    \multirow{3}{*}{\shortstack[l]{\textbf{Spatial}\\\textbf{Grounding}}} 
      & Visual Consistency & 0.87 & 0.67 & 0.60 & 0.60 & \textbf{1.00} \\
      & Dynamic Feasibility & 0.80 & \textbf{1.00} & 0.80 & \textbf{1.00} & \textbf{1.00} \\
      & Task Completion & 0.40 & 0.67 & 0.00 & 0.00 & \textbf{0.93} \\
    \midrule

    \multirow{3}{*}{\shortstack[l]{\textbf{Language}\\\textbf{Control}}} 
      & Visual Consistency & 0.60 & 0.80 & 0.53 & \textbf{1.00} & 0.93 \\
      & Dynamic Feasibility & 0.93 & 0.93 & \textbf{1.00} & \textbf{1.00} & 0.93 \\
      & Task Completion & 0.13 & 0.07 & 0.00 & 0.00 & \textbf{0.80} \\
    \midrule

    \multirow{3}{*}{\shortstack[l]{\textbf{Scene}\\\textbf{Reasoning}}} 
      & Visual Consistency & \textbf{1.00} & 0.47 & 0.60 & 0.93 & \textbf{1.00} \\
      & Dynamic Feasibility & \textbf{1.00} & \textbf{1.00} & 0.92 & \textbf{1.00} & \textbf{1.00} \\
      & Task Completion & 0.93 & 0.37 & 0.20 & 0.33 & \textbf{0.87} \\
    \midrule
    
    \multirow{3}{*}{\textbf{Average}} 
      & Visual Consistency & 0.77 & 0.67 & 0.53 & 0.91 & \textbf{0.99} \\
      & Dynamic Feasibility & 0.93 & 0.97 & 0.94 & \textbf{1.00} & 0.99 \\
      & Task Completion & 0.48 & 0.45 & 0.13 & 0.13 & \textbf{0.84} \\
    \bottomrule
  \end{tabular}
  \vspace{3mm}
    \caption{\textbf{Quantitative comparison of video models across diverse 3D navigation tasks.} We evaluate open-source and closed-source models across five specialized categories based on three core metrics: Visual Consistency, Dynamic Feasibility, and Task Completion. Each reported score represents the mean performance over five independent sampling trials, with the best results highlighted in bold.}
  \vspace{-2mm}
  \label{tab:worldmodelbenchmark}
\end{table}

\subsection{Evaluating Video Models}
\label{sec:worldmodelbenchmark}

Ensuring physical plausibility and instruction alignment is critical for video-based navigation. However, models vary in their understanding of different tasks due to discrepancies in training data and architectures. To evaluate this, we designed 15 distinct tasks covering five dimensions: object navigation, precise navigation, spatial grounding, language control and scene reasoning. We benchmarked several open-source models, including Wan 2.2 I2V 14B \cite{wan}, HunyuanVideo-1.5 I2V 8B \cite{huanyuan}, Cosmos-Predict 2.5 14B \cite{cosmos}, and LVP \cite{lvp}, alongside the closed-source Wan 2.6 \footnote{https://tongyi.aliyun.com/wan/}.

To quantify performance, we established a navigation metric focused on the following three aspects:

(1) Visual Consistency: This evaluates whether the scene remains consistent during navigation, including static background objects moving or rotating, and the hallucination of non-existent objects

(2) Dynamic Feasibility: This assesses whether the camera motion adheres to dynamic constraints.

(3) Task Completion: This evaluates whether the generated video aligns with the instruction's intent.

 \vspace{1mm}

To ensure evaluation robustness, we perform five independent samplings for each model per task. While Qwen3-VL excels at relative assessment for selecting optimal candidates, it lacks the necessary calibration for isolated absolute scoring. Therefore, to maintain objective and accurate benchmarking, we invited experienced UAV pilots to score the synthesized videos, ensuring the results precisely reflect real-world navigation quality.

As shown in Table \ref{tab:worldmodelbenchmark}, Wan 2.6 achieves the highest overall performance across nearly all metrics. Notably, HunyuanVideo-1.5 exhibits specialized proficiency in object navigation and spatial grounding, surpassing Wan 2.2 in these target-centric dimensions. However, it falls considerably behind Wan 2.2 in scene reasoning tasks, suggesting a more constrained capacity for scene understanding. Regarding Cosmos 2.5, while it demonstrates basic competence in simple object navigation, its performance drops significantly in rare actions. In these instances, the model frequently suffers from severe mode collapse. For example, when tasked with traversing a circular frame, Cosmos 2.5 not only fails to complete the objective but also produces videos with violent fluctuations in stability. While LVP gets high scores in consistency and feasibility, this is because it stays still to keep the video stable, which accounts for its low success rate. We attribute this behavior to its training on manipulation datasets, which lack the significant ego-motion priors required for 3D navigation tasks.  Additionally, our qualitative evaluation identified four primary failure modes that reflect the current limitations of video-based world models in 3D navigation, which are discussed in detail in the Appendix \ref{sec:pattern_bias}.

\subsection{Real-World Results}
\textbf{Real-world deployment:} The platform is equipped with an Intel RealSense camera for visual perception and LiDAR for pose estimation. The task image and prompt are sent to Wan 2.6 video model via an API. Once the video is generated, an action decoder (deployed on a laptop) extracts the waypoint sequence. The low-level action module then plans a trajectory in real time based on these waypoints and the current state estimated by the Fast-LIVO2 \cite{fastlivo2}. This estimation process utilizes LiDAR and IMU data to provide accurate positioning. Both the low-level planning and the pose estimation modules are deployed on a Jetson Orin NX for efficient on-board processing.

\textbf{Surprising Generalization:} As shown in Fig.~\ref{fig:toutu}, we evaluated the video model across a wide range of tasks and scenes, spanning from indoor to outdoor environments. These scenarios include object-based navigation, common-sense reasoning, 3D spatial flight, and semantic scene understanding. Notably, many of these tasks are inherently difficult to describe via language or require exhaustive rule-based designs in traditional frameworks. For instance, to observe an object occluded behind an obstacle, the agent must bypass the obstruction while performing smooth, natural yaw rotations to maintain a continuous line of sight. While traditional methods rely on complex hand-coded rules for view-space calculations and feedback control, our approach manages these coordinated behaviors intuitively. Benefiting from internet-scale training data, the model demonstrates remarkable zero-shot generalization. This effectively validates our pipeline’s capability in scene understanding, spatial awareness, and complex behavior execution, allowing implicit action intents to be efficiently transferred into real-world robot movements.

\textbf{Emergent Perception in Dreaming:} Beyond navigation, video models exhibit emergent capabilities in handling low-level perception challenges. Our experiments show that the model can zero-shot restore high-definition scenes from degraded inputs. Furthermore, it demonstrates the ability to expand the observation space through dreaming. Detailed results are presented in Appendix \ref{sec:emergent}.

\subsection{Ablation Studies}
\label{sec:albation}

\begin{figure}[h]
  \centering
  \includegraphics[width=0.98\linewidth]{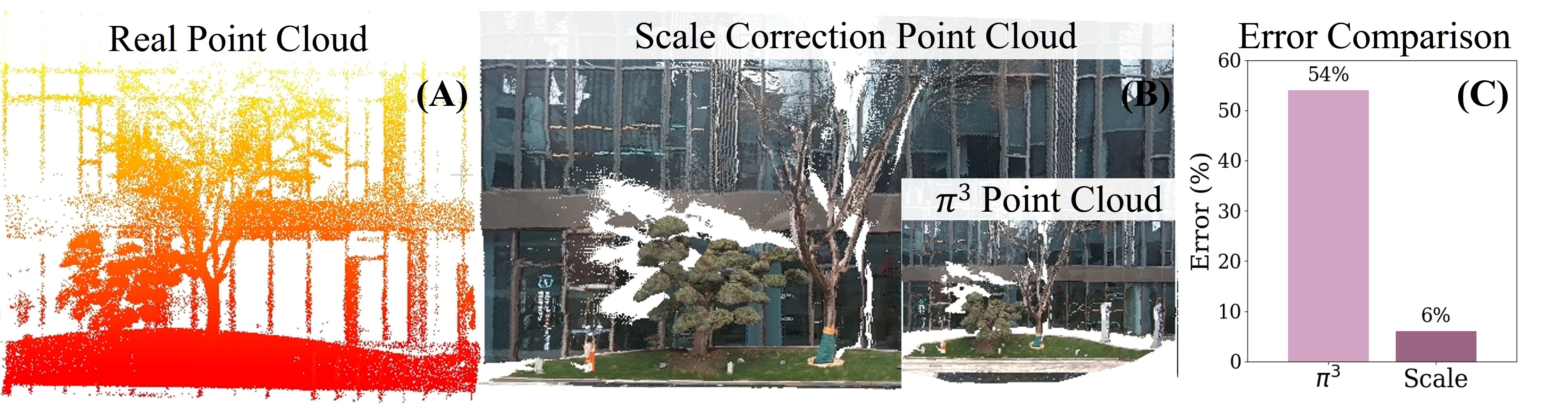}
    \caption{\textbf{Visualization and Quantitative Analysis of Scale Correction.} (A) Reference ground-truth point cloud captured from the real-world environment. (B) Comparative visualization of reconstructed point clouds before and after metric scale alignment. (C) The error comparison of the methods. \textbf{Note: The relative picture size of (A) and (B) are rendered to be the same as their true physical proportions.}}
  \label{fig:scale}
\end{figure}

\textbf{Ablation Study on Performance Components:} We first evaluate the effectiveness of the absolute physical scale estimation within the action decoder. We test several outdoor tasks characterized by significant translational displacements. For example, as illustrated in Fig.~\ref{fig:scale}, the original physical scale decoded by $\pi^3$ is only approximately 46\% of the ground truth. Specifically, the relative scale error is defined as $\mathcal{E}_{rel} = \frac{|\mathcal{S} - \mathcal{S}_{gt}|}{\mathcal{S}_{gt}}$, where $\mathcal{S}$ and $\mathcal{S}_{gt}$ denote the estimated and ground-truth scale factors, respectively. This discrepancy occurs because the model may fail to find reasonable implicit references in unseen environments, leading to the failure of absolute scale estimation in such outdoor scenes. However, with our proposed scale correction, we can reduce the relative error to the 10\% level, which significantly enhances navigation reliability in open outdoor scenarios.

\begin{figure}[h]
  \centering
  \small
  \renewcommand{\arraystretch}{1.6} 
  
  \begin{minipage}[b]{0.62\textwidth}
    \centering
    \setlength{\tabcolsep}{4pt} 
    \begin{tabular}{@{}lccc@{}}
      \toprule
      \textbf{Strategy} & \makecell[c]{Visual \\ Consistency} & \makecell[c]{Dynamic \\ Feasibility} & \makecell[c]{Task \\ Completion} \\ \midrule
      (1) Simple     & 0.80 & 0.80 & 0.41 \\
      (2) Kinematic  & 0.83 & 0.86 & 0.46 \\
      (3) Decomposed & \textbf{0.83} & \textbf{0.86} & \textbf{0.73} \\
      (4) Rewritten  & 0.83 & 0.80 & 0.58 \\ \bottomrule
    \end{tabular}
    \vspace{2mm} 
    \captionof{table}{\textbf{Performance Evaluation Across Different Prompting Strategies.}}
    \label{tab:prompt_ablation}
  \end{minipage}
  \hfill
  \begin{minipage}[b]{0.35\textwidth}
    \centering
    \includegraphics[width=\linewidth]{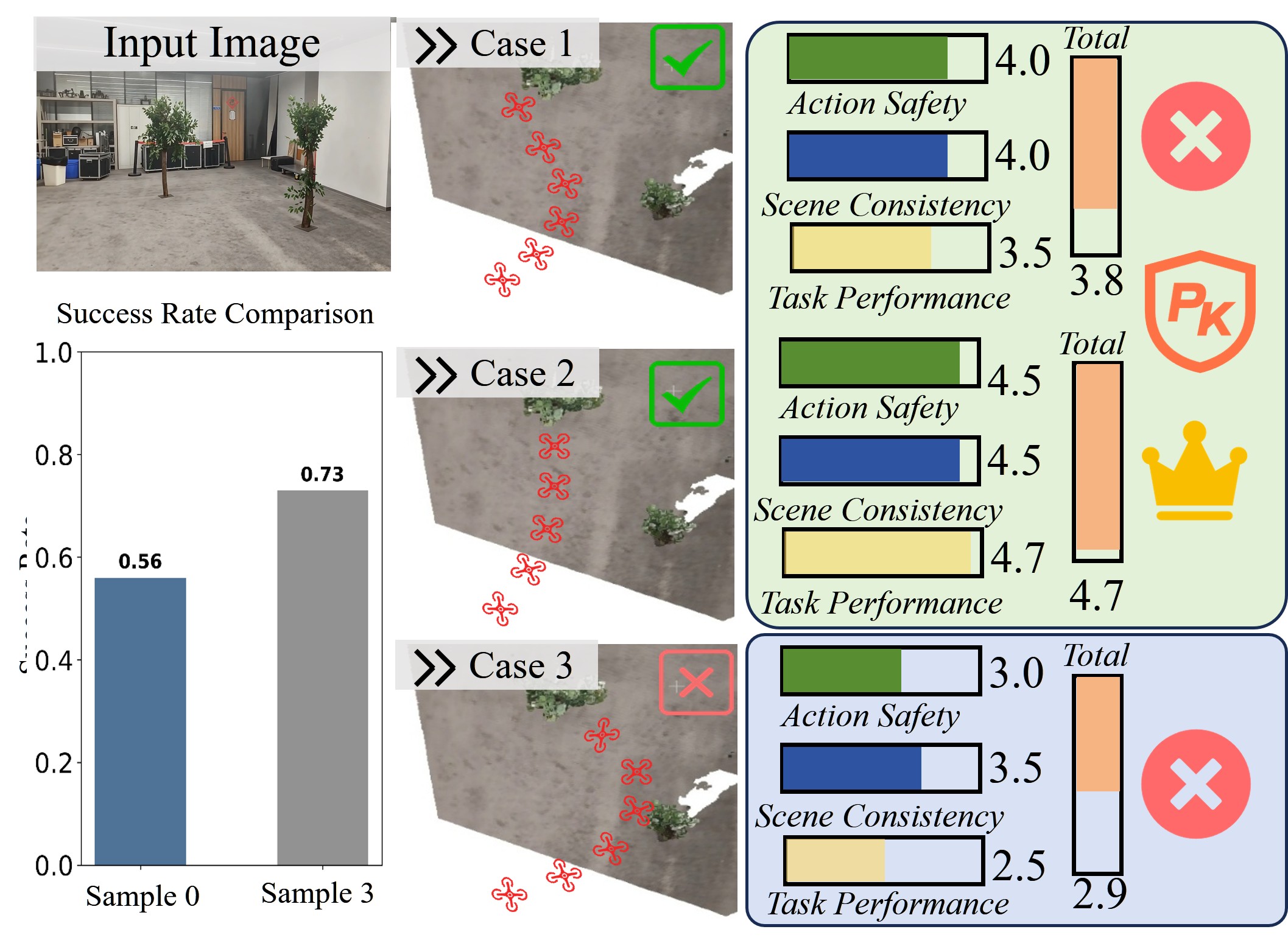}
    \vspace{0mm} 
    \captionof{figure}{\textbf{The ablation experiment of sampling-based optimization.}}
    \label{fig:sampling_ablation}
  \end{minipage}
\end{figure}

Furthermore, we investigate the impact of "dreaming before planning". We vary stochastic seeds across multiple sampling trials to analyze the correlation between sampling budget and success rate. As shown in Fig.~\ref{fig:sampling_ablation}, increasing the sampling budget significantly improves the task success rate. This highlights a fundamental advantage of video-based foundation models in robotics: the ability to predict multiple future visual states allows the system to proactively bypass unsafe behaviors and select high-value actions through VLM-based reward evaluation.



\textbf{Prompt design in 3D navigation:} Compared to LLMs, video generation models exhibit higher sensitivity to the design of prompts. We selected six challenging tasks and sampled each task five times using Wan2.6 as the base model. We categorized the prompt descriptions into four hierarchical levels of detail: (1) \textit{Simple}: basic task descriptions; (2) \textit{Dynamic-Aware}: simple descriptions augmented with physical and kinematic constraints; (3) \textit{Decomposed}: highly detailed constraints that decompose motion into specific sub-events with precise timing; and (4) \textit{Prompt-Rewritten}: the simple descriptions processed through an automated prompt rewriting interface. As shown in Table \ref{tab:prompt_ablation}, the second setting outperforms the first by 6\% in dynamic feasibility and enhances the task success rate by 5\%. While the third setting maintains levels of visual consistency and dynamic feasibility similar to the second, it achieves a significant 27\% improvement in task success rate. We attribute this significant improvement to the detailed action decomposition. Interestingly, the results for the \textit{Prompt-Rewritten} setting show a notable improvement over settings (1) and (2) in terms of success rate. However, it suffers a 16\% decrease compared to the \textit{Decomposed} setting. This discrepancy arises because when the initial prompt lacks sufficient detail, the automated rewriting process tends to introduce semantic ambiguity. This can lead to misinterpretations of sophisticated behaviors, resulting in mission failure. These findings demonstrate that providing detailed kinematic constraints and explicit intent explanations is key to adapting video models for your own robotic platforms.


\section{Conclusion}
\label{sec:conclusion}

We introduce NavDreamer, a framework leveraging generative video models for zero-shot 3D navigation. By employing VLM-based trajectory selection and an inverse dynamics model with metric depth priors, we ensure reliable, scale-accurate navigation. We also propose a comprehensive benchmark spanning five dimensions. Experiments demonstrate robust generalization to unseen environments, successfully bridging visual imagination and physical action.

\clearpage

\section{Limitation}
\label{sec:Limitation}
Although our framework demonstrates impressive zero-shot 3D navigation capabilities, it encounters significant challenges when executing tasks requiring high agility and precision—such as aggressive aerobatic maneuvers or traversing narrow hoops. Furthermore, the substantial computational latency remains a critical bottleneck, as current video models typically require one to two minutes to synthesize a complete sequence via API, thereby limiting performance in high-reactivity scenarios that demand immediate feedback. To address these limitations, our future work will focus on fine-tuning generative video models on specialized aerial robot datasets, while simultaneously investigating quantization and model compression techniques to achieve the lightweight inference necessary for real-time 3D navigation.



\bibliography{example}  

\clearpage 
\appendix
\AppendixTableSetup
\section{Overview}
In this supplementary material, we present additional details and results to complement the main manuscript.
Section B.1 discusses the four primary failure modes of video models in 3D navigation, specifically analyzing Pattern Bias, Reward Hacking, Mode Collapse, and Instruction Neglect. Section B.2 outlines the task specifications and benchmark descriptions for our 3D Drone Navigation Benchmark. Section B.3 details our experiments on robust perception, including blind deblurring, low-light enhancement, and super-resolution within the navigation pipeline. Section B.4 presents the prompt design ablation experiments, providing the full hierarchy of prompts (Simple, Refined, and Detailed) for various navigation tasks. Section B.5 details the structured VLM prompt design. Finally, Section B.6 directs the reader to extensive video results and real-world flight performance, showcasing multi-perspective visualizations of our zero-shot navigation framework.


\section{Experiment Details}

\begin{figure}[H]
  \centering
  \includegraphics[width=0.98\linewidth]{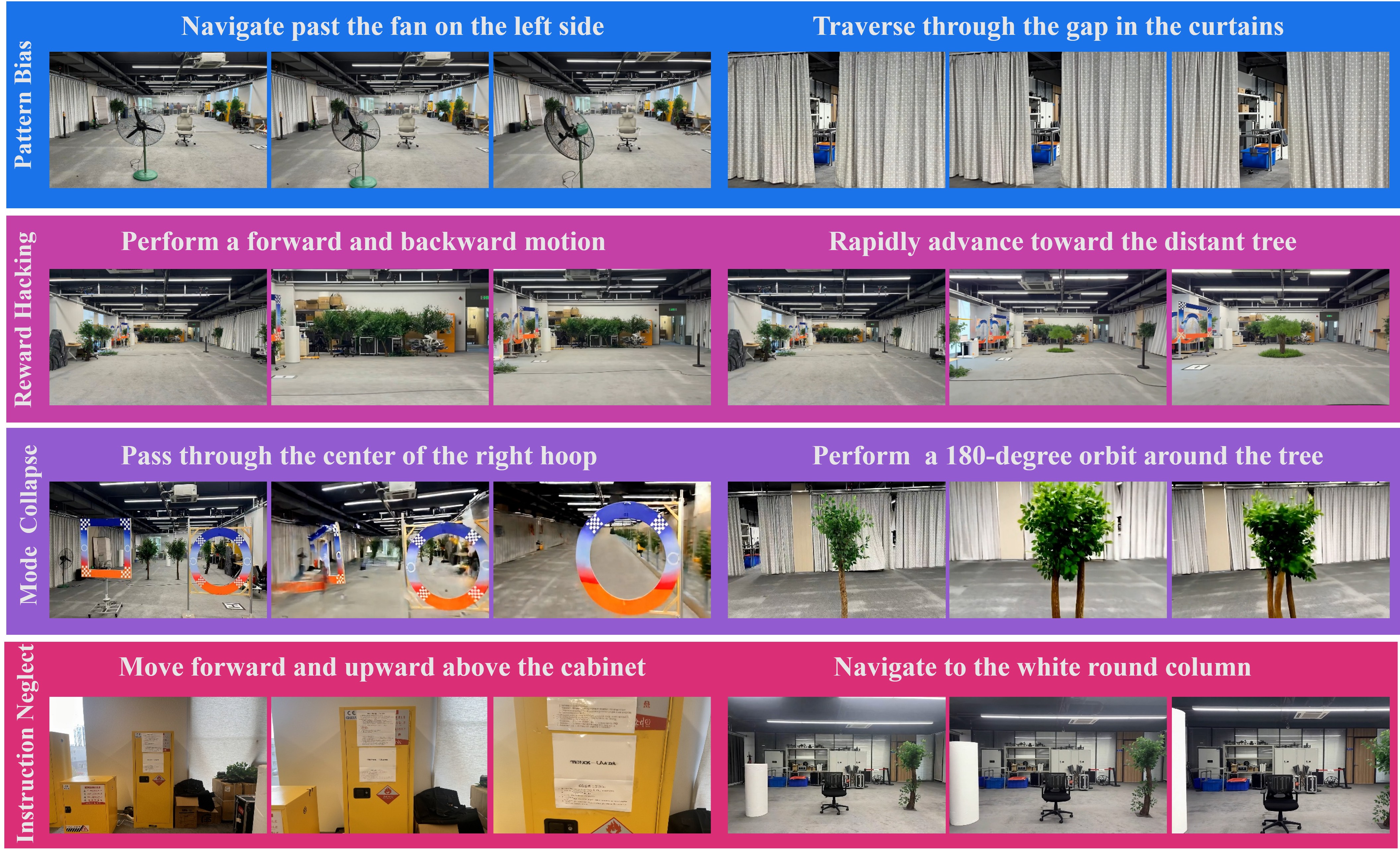}
  \caption{Four main failure modes in 3D navigation.}
  \label{fig:failure}
\end{figure}

\subsection{Failure Modes of Video Models in 3D navigation}
\label{sec:pattern_bias}
\textbf{Pattern Bias}: Just like LLMs inherit and amplify social stereotypes from uncurated internet data \cite{bias}, many video models also exhibit strong pattern biases. For example, as shown in the first column of Fig.~\ref{fig:failure}, models often assume that an electric fan is rotating. We suggest this happens because most fans in the training datasets are shown moving, so the model follows this common pattern regardless of whether the fan in our specific scene is actually powered. Similarly, we observed the stationary curtains in a room without any physical triggers often spontaneously part and slide toward both sides. These cases show that the models rely more on common data patterns than on the actual logic of the current scene.

\textbf{Reward Hacking}: During testing, many models exhibit "deceptive" strategies to satisfy the prompt's visual requirements without performing the real physical movement. For example, in the 'fast move' task, the camera is required to reach a tree at the far end of the scene within five seconds. If the generated motion is too slow to reach the target on time due to an error, the model may simply hallucinate a new tree directly in front of the camera to satisfy the prompt. Alternatively, models sometimes manipulate the focal length by using a zoom-out effect to create a "looming" visual flow. This mimics forward progress even though the camera remains spatially static. These results show that current video models often adopt seemingly correct behaviors to satisfy the prompt requirements, which leads to reward hacking.

\textbf{Mode Collapse}: When tasked with rare actions or scenarios that involve significant environmental transitions, video models often suffer from severe instabilities, a phenomenon particularly evident in Cosmos 2.5. As shown in the third column of Fig.~\ref{fig:failure}, when traversing a circular hoop, the environment becomes fragmented and unstable. Similarly, during an orbital flight around a tree, the model fails to maintain object integrity, resulting in the hallucination of multiple redundant branches. These violent fluctuations indicate a breakdown in the model's ability to maintain spatial and temporal coherence under complex dynamics.

\textbf{Instruction Neglect}: As depicted in the fourth column of Fig.~\ref{fig:failure}, models occasionally disregard specific constraints or commands within the prompt. For example, when instructed to ascend above a cabinet and navigate toward a white column, the generated video fails to execute the requested vertical motion. While this behavior is more prevalent in complex tasks due to linguistic ambiguity or limited reasoning capacity, it also manifests in simpler scenarios despite using consistent seeds and prompts but is effectively mitigated by our sampling-based optimization.

\subsection{Task Setup and Description in 3D Navigation Benchmark}

\begin{table}[H]
\centering
\small
\begin{tabular}{@{}lll@{}}
\toprule
\textbf{Task Type} & \textbf{Task Name} & \textbf{Task Description} \\ \midrule
\multirow{3}{*}{Object Navigation} & Find Chair & Navigate to the black chair and stop directly in front. \\
 & Find Column & Navigate to the white round column and stop directly in front. \\
 & Find Tree & Navigate to the green tree and stop directly in front. \\ \midrule
\multirow{3}{*}{Precise Navigation} & Above Cabinet & Fly forward and upward to stop 0.5m above the cabinet center. \\
 & Behind Rock & Orbit the large rock to its rear while keeping gaze on the target. \\
 & Left of Tree & Move left past the tree until it is completely out of the frame. \\ \midrule
\multirow{3}{*}{Spatial Grounding} & Circle Orbit & Perform a 180-degree orbit around the tree. \\
 & Gate Traversal & Pass through the center of the circular hoop to the space beyond. \\
 & Midline Stop & Advance and stop on the line connecting the pillar and the tree. \\ \midrule
\multirow{3}{*}{Language Control} & Timed Stop & Accelerate for 3s toward the tree, then hover stationary for 2s. \\
 & Round Trip & Approach the plants and retrace the path back to the starting position. \\
 & Turn and Advance & Rotate 45$^\circ$ to the right and fly straight forward across the room. \\ \midrule
\multirow{3}{*}{Scene Reasoning} & Find Kitchen & Identify scene features and navigate to the room suitable for cooking. \\
 & Find Exit & Observe environment markers to locate and navigate toward the exit. \\
 & Find Bathroom & Identify scene features and navigate to the room for showering. \\ \bottomrule
\end{tabular}
\vspace{3mm}
\caption{Benchmark Task Suite for 3D Drone Navigation}
\label{tab:task_design}
\end{table}

\subsection{Emergent Perception in Dreaming}
\label{sec:emergent}

\begin{figure}[H]
  \includegraphics[width=0.98\linewidth]{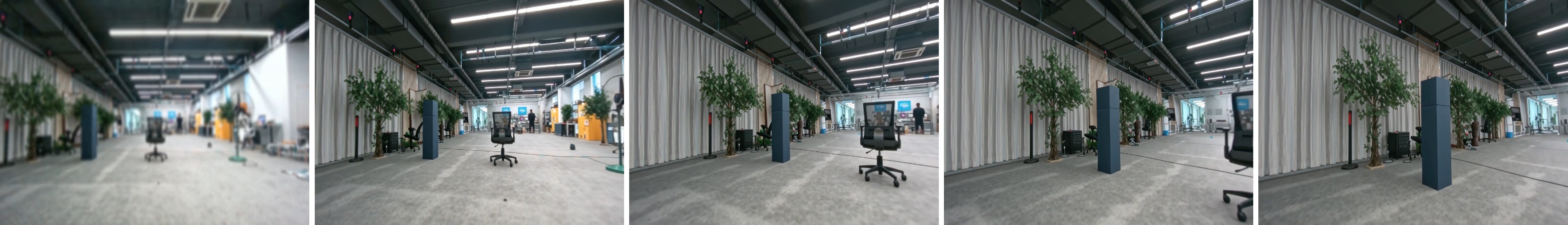}
  \caption{\textbf{Blind deblurring:} Generate a first-person perspective. First, unblur the image, including the background. Then, move to the blue cuboid in the scene smoothly and steadily, stopping at a safe distance directly in front of it."}
  \label{fig:blur}
\end{figure}
\begin{figure}[H]
  \includegraphics[width=0.98\linewidth]{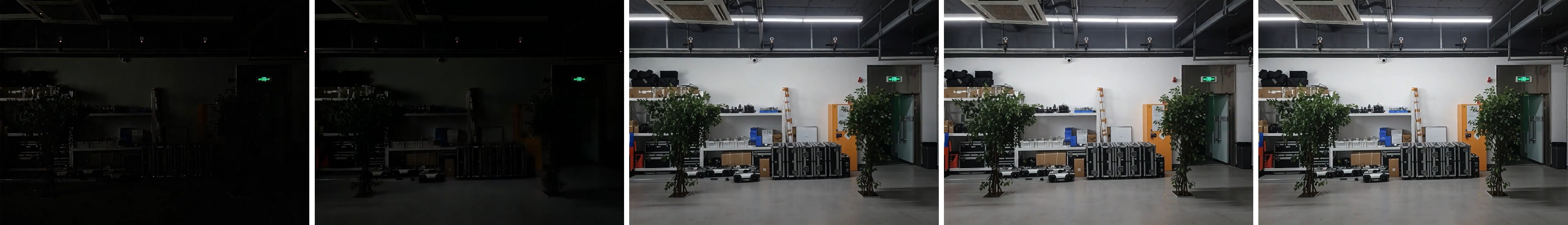}
  \caption{\textbf{Low-light enhancing:} Fully restore the light in this image. Static camera perspective, no zoom or pan.”}
  \label{fig:light}
\end{figure}
\begin{figure}[H]
  \includegraphics[width=0.98\linewidth]{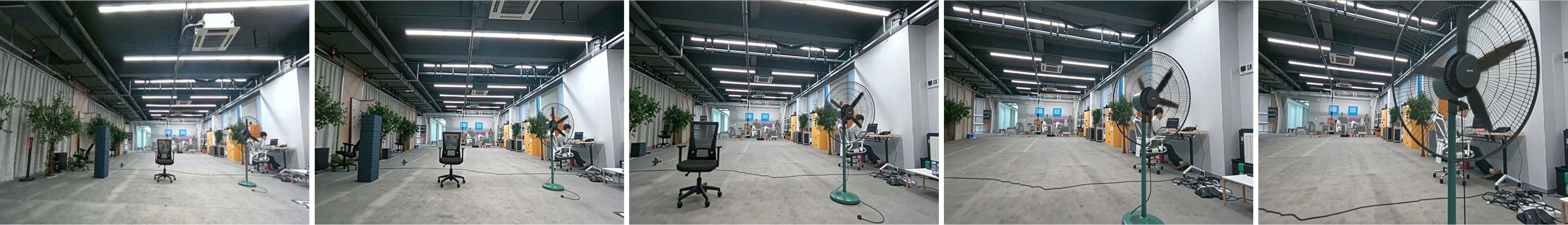}
  \caption{\textbf{Super-resolution:} Generate a first-person perspective. First, perform super-resolution on this image. Then, move to the electric fan in the scene smoothly and steadily, stopping at a safe distance directly in front of it.}
  \label{fig:resolution}
\end{figure}
\begin{figure}[H]
  \includegraphics[width=0.98\linewidth]{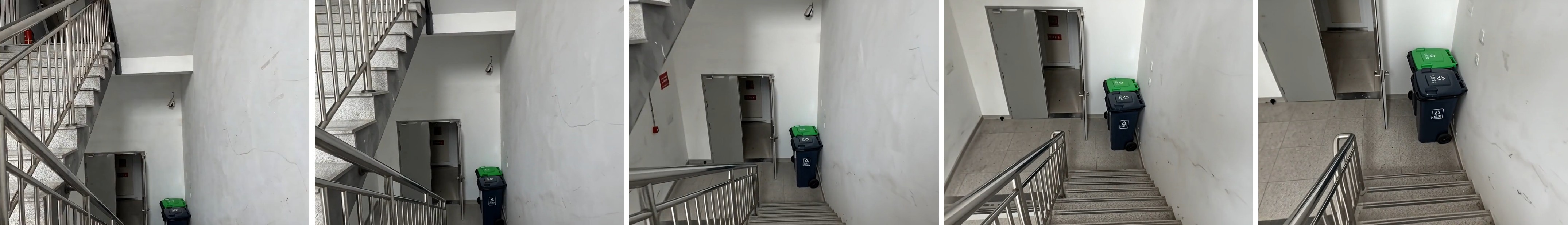}
  \caption{\textbf{Downstairs:} Generate a first-person perspective video. Starting at the top of the stairs, move forward with steady steps, taking each step quickly and sequentially, maintaining a consistent downward trajectory.  The video must end precisely at the completion of the last step, stopping at the trash can at the bottom of the stairs.}
  \label{fig:downstairs}
\end{figure}

Video models demonstrate a unified ability to handle specialized computer vision tasks. For instance, \cite{goolevideo} demonstrated solutions across diverse capabilities ranging from perception and modeling to manipulation and reasoning. We aim to investigate whether these dreaming abilities can be extended to benefit 3D navigation tasks in perception. To this end, we designed experiments focusing on three critical perception challenges encountered during drone flight: low-resolution (Fig. \ref{fig:resolution}), motion blur (Fig. \ref{fig:blur}), and low-light conditions (Fig. \ref{fig:light}). The video model demonstrates a remarkable ability to generalize across tasks that previously required specialized Computer Vision expert models. Specifically, it can reconstruct reliable, high-definition scenes from blurred, low-resolution, or poorly lit inputs. This restoration of visual clarity significantly enhances the robustness and reliability of drone navigation in complex environments.
 
Furthermore, we observed an emergent capability during the "going downstairs" task (Fig. \ref{fig:downstairs}), where the model proactively tilts the synthesized view downward to imagine the necessary observation space beyond the initial observation of the scene limited by the camera's FoV . This dreaming paradigm demonstrate that video models can reasonably enumerate future environment observation when prompted with a partial snapshot of the environment. The predictive capabilities and 3D understanding from the video model is a potential solution to the partially observable nature of navigation tasks.  


\subsection{Prompt Design Ablation Experiment}
\begin{table}[H]
\centering

\small
\renewcommand{\arraystretch}{1.2} 
\begin{tabularx}{\textwidth}{@{}l l X@{}}
\toprule
\textbf{Target Task} & \textbf{Prompt Level} & \textbf{Prompt Content (Input to Generative Model)} \\ \midrule

\multirow{3}{*}{\textbf{Circle Tree}} 
 & (1) Simple & A first-person perspective video of a 360-degree circular orbit around a green tree centered in the frame. \\ \cmidrule(l){2-3}
 & (2) Refined & First-person perspective. The camera performs a smooth and steady 360-degree circular orbit around the green tree at a constant speed and fixed height, maintaining a strictly eye-level view without any vertical bobbing, tilting, or camera shake. \\ \cmidrule(l){2-3}
 & (3) Detailed & First-person perspective. The camera performs a fast and steady 360-degree circular orbit around the green tree. The rotation completes a full 360-degree circle, ensuring the camera ends exactly back at the starting position and orientation. All environmental details, including the grey floor and patterned curtains, must remain perfectly consistent. \\ \midrule

\end{tabularx}

\label{tab:appendix_prompt_ablation}
\end{table}

\begin{table}
\centering
\small
\renewcommand{\arraystretch}{1.2} 
\begin{tabularx}{\textwidth}{@{}l l X@{}}
\toprule
\textbf{Target Task} & \textbf{Prompt Level} & \textbf{Prompt Content (Input to Generative Model)} \\ \midrule

\multirow{3}{*}{\textbf{Behind Rock}} 
 & (1) Simple & A first-person perspective video circling around the large black rock to reach the area behind it. \\ \cmidrule(l){2-3}
 & (2) Refined & First-person perspective. The camera performs a smooth and steady circular movement to reach the rear of the large black rock, maintaining a constant eye-level view without any vertical bobbing or camera shake throughout the trajectory. \\ \cmidrule(l){2-3}
 & (3) Detailed & First-person perspective. The camera performs a smooth, steady circular motion around the large black rock to reach its rear. Crucially, the camera's gaze remains fixed on the specific area behind the rock at all times. The entire indoor studio environment must remain perfectly consistent and static. \\ \midrule

\multirow{3}{*}{\textbf{Find Kitchen}} 
 & (1) Simple & Move to the room where cooking is possible. \\ \cmidrule(l){2-3}
 & (2) Refined & First-person perspective. Perform a smooth and steady gliding motion at a constant height to enter the room where cooking is possible. The movement must be fluid with zero camera shake or erratic tilting. \\ \cmidrule(l){2-3}
 & (3) Detailed & First-person perspective. Carefully analyze the visual landmarks and functional attributes of the two distinct spaces. Based on this semantic reasoning, execute a smooth transition into the specific area designed for food preparation while maintaining absolute environmental consistency. \\ \midrule

\multirow{3}{*}{\textbf{Fast Move}} 
 & (1) Simple & First-person perspective video where the camera moves very fast toward the green tree in the distance and stops right in front of it. \\ \cmidrule(l){2-3}
 & (2) Refined & First-person perspective. The camera performs a smooth, high-speed forward movement toward the green tree, accelerating continuously to create a dynamic sense of depth, followed by a precise and stable stop directly in front of the tree. \\ \cmidrule(l){2-3}
 & (3) Detailed & In the first 3 seconds, the camera rapidly moves forward toward the green tree, accelerating continuously to create motion blur on background elements. At the exact end of the third second, the camera comes to an immediate, precise stop at a one-meter distance. For the remaining 2 seconds, the camera remains completely stationary. \\ \midrule

\multirow{3}{*}{\textbf{Round Trip}} 
 & (1) Simple & A first-person perspective video where the camera moves forward toward the green plants and then moves backward to the starting position. \\ \cmidrule(l){2-3}
 & (2) Refined & First-person perspective. The camera moves smoothly forward toward the plants along the center axis. At the midpoint of the path, the camera comes to a complete stop, then pulls back at the same constant speed until it returns to the start. \\ \cmidrule(l){2-3}
 & (3) Detailed & The camera approaches the plants from the center axis, in a first-person perspective. When it reaches the midpoint, the camera stops. Then, the camera pulls back at the same speed, retracing the original path completely. The last frame of the video is exactly the same as the first frame. \\ \midrule
\multirow{3}{*}{\textbf{Turn and Advance}} 
 & (1) Simple & A first-person perspective video where the camera turns 45 degrees to the right and then moves straight forward across the open studio. \\ \cmidrule(l){2-3}
 & (2) Refined & First-person perspective. The camera performs a smooth 45-degree right turn, followed by a steady and fluid forward movement. The transition from rotation to translation must be seamless, maintaining constant height without unintended rotation. \\ \cmidrule(l){2-3}
 & (3) Detailed & In the first 1 second, the camera smoothly rotates 45 degrees to the right. Immediately after finishing the rotation, it moves straight forward across the studio without any further rotation. All environmental details—including overhead lights and rigging—must remain perfectly consistent and static. \\ \bottomrule
\end{tabularx}
\vspace{3mm}
\caption{Detailed Comparison of Prompt Hierarchies across Navigation Tasks}
\end{table}

\clearpage

\definecolor{promptbgcolor}{RGB}{245,245,245} 
\definecolor{promptframecolor}{RGB}{139,69,19} 
\definecolor{prompttitlebg}{RGB}{139,69,19}   
\subsection{VLM Prompt Design}
\begin{tcolorbox}[
    colback=promptbgcolor,
    colframe=promptframecolor,
    coltitle=white,
    title=\textbf{Prompt for Video Quality Ranking and Task Verification},
    fonttitle=\bfseries\large,
    boxrule=1.0pt,
    arc=2mm,
    left=2mm, right=2mm, top=2mm, bottom=2mm,
    breakable,        
    before skip=2mm,  
    after skip=2mm    
]

\small 
\textbf{Role Definition:} You are a drone video quality ranking arbitrator and task verification expert. You have \texttt{\{num\_videos\}} candidate videos, all based on the same starting scene and instruction.

\vspace{1mm}
\textbf{Core Task:} Verify each video's success status individually, then evaluate them together to determine a weighted score and final rank.

\par\noindent\rule{\textwidth}{0.4pt} 

\textbf{Step 1: Task Verification (Individual Check -- Determine Status)} \\
Analyze the logical structure and verify against strict red lines.
\begin{itemize}[leftmargin=*, noitemsep, topsep=0pt]
    \item \textbf{Structure Analysis}:
    \begin{itemize}[leftmargin=3mm]
        \item \textit{Type A (Single-Stage)}: Continuous action or final state (e.g., ``Fly forward'', ``Hover'').
        \item \textit{Type B (Multi-Stage)}: Sequential logic (e.g., ``First A then B'', ``Next...'').
    \end{itemize}
    \item \textbf{Disqualification Criteria (Status = Fail)}:
    \begin{itemize}[leftmargin=3mm]
        \item \textit{Motion Failure}: Instruction implies movement, but view barely changes.
        \item \textit{Directional Deviation}: Movement is opposite to instruction.
        \item \textit{Target Miss}: Stopping far from the required destination.
        \item \textit{Trajectory Incompleteness}: Executing only a fraction of a required shape.
        \item \textit{Stage Omission (Type B)}: Skipping intermediate steps (shortcutting).
        \item \textit{Sequence Error (Type B)}: Wrong order of actions.
        \item \textit{General Failures}: Hallucinations, Physics Violation (teleportation), Obscuring Blur.
    \end{itemize}
    \item \textbf{Passing Criteria (Status = Pass)}: Must NOT violate red lines. Multi-Stage tasks must demonstrate the intent of \textbf{all} main stages.
\end{itemize}

\vspace{1mm}
\textbf{Step 2: Ranking Evaluation Framework (Relative Check)} \\
Evaluate all videos on the following dimensions:
\begin{enumerate}[leftmargin=*, noitemsep, topsep=0pt]
    \item \textbf{Task Performance (TP) (Highest Priority)}:
    \begin{itemize}[leftmargin=3mm]
        \item \textit{Logical Fidelity}: For Multi-Stage, prioritize coverage over perfection. A blurry ``Pass'' outranks a shortcut ``Fail''.
        \item \textit{Execution Quality}: Magnitude (Significant $>$ Minimal), Target Accuracy (Correct $>$ Incorrect), Trajectory Correctness (Correct $>$ Distorted).
    \end{itemize}
    \item \textbf{Action Safety (AS)}: Physics \& Stability (Coherent motion $>$ Teleportation; Smooth control $>$ Jitter).
    \item \textbf{Scene Consistency (SC)}: Visual Integrity (Stable environment $>$ Hallucinations/Clipping).
\end{enumerate}

\vspace{1mm}
\textbf{Step 3: Scoring Standards \& Ranking} \\
Assign scores (0.0 - 5.0) based on the levels below. Note that a "Fail" status usually corresponds to lower levels:
\begin{itemize}[leftmargin=*, noitemsep, topsep=0pt]
    \item \textbf{Level 1 (Failure, 0.0--1.0)}: Violates core logic. Stationary, wrong direction. (\textbf{Status: Fail})
    \item \textbf{Level 2 (Poor, 1.0--2.9)}: Attempted but failed significantly. (\textbf{Status: Fail/Pass})
    \item \textbf{Level 3 (Average, 3.0--3.9)}: Completed but lacks polish. Shaky path. (\textbf{Status: Pass})
    \item \textbf{Level 4 (Good, 4.0--4.5)}: Precise requirements. Good trajectory. (\textbf{Status: Pass})
    \item \textbf{Level 5 (Excellent, 4.8--5.0)}: Flawless execution. Natural physics. (\textbf{Status: Pass})
\end{itemize}

\textbf{Final Calculation}:
\texttt{Total Score = (TP * 1.4 + AS * 0.8 + SC * 0.8) / 3}. Rank descending.

\par\noindent\rule{\textwidth}{0.4pt} 

\textbf{Input Information:} \\
\textbf{Current Instruction:} \texttt{\{instruction\}}

\textbf{Output Format} (Strictly follow): \\
\texttt{Video 1: <score> X.X </score> | Status: [Pass/Fail] | TP: X.X | AS: X.X | SC: X.X | Reason: [Concise explanation]} \\
... \\
\texttt{Best: Video N}

\end{tcolorbox}
\vfill
\clearpage

\subsection{Video Results and Real-World Results}

\captionsetup[subfigure]{labelformat=empty}
\begin{figure}[H] 
    \centering
    \begin{subfigure}[b]{0.9\textwidth}
        \includegraphics[width=\textwidth]{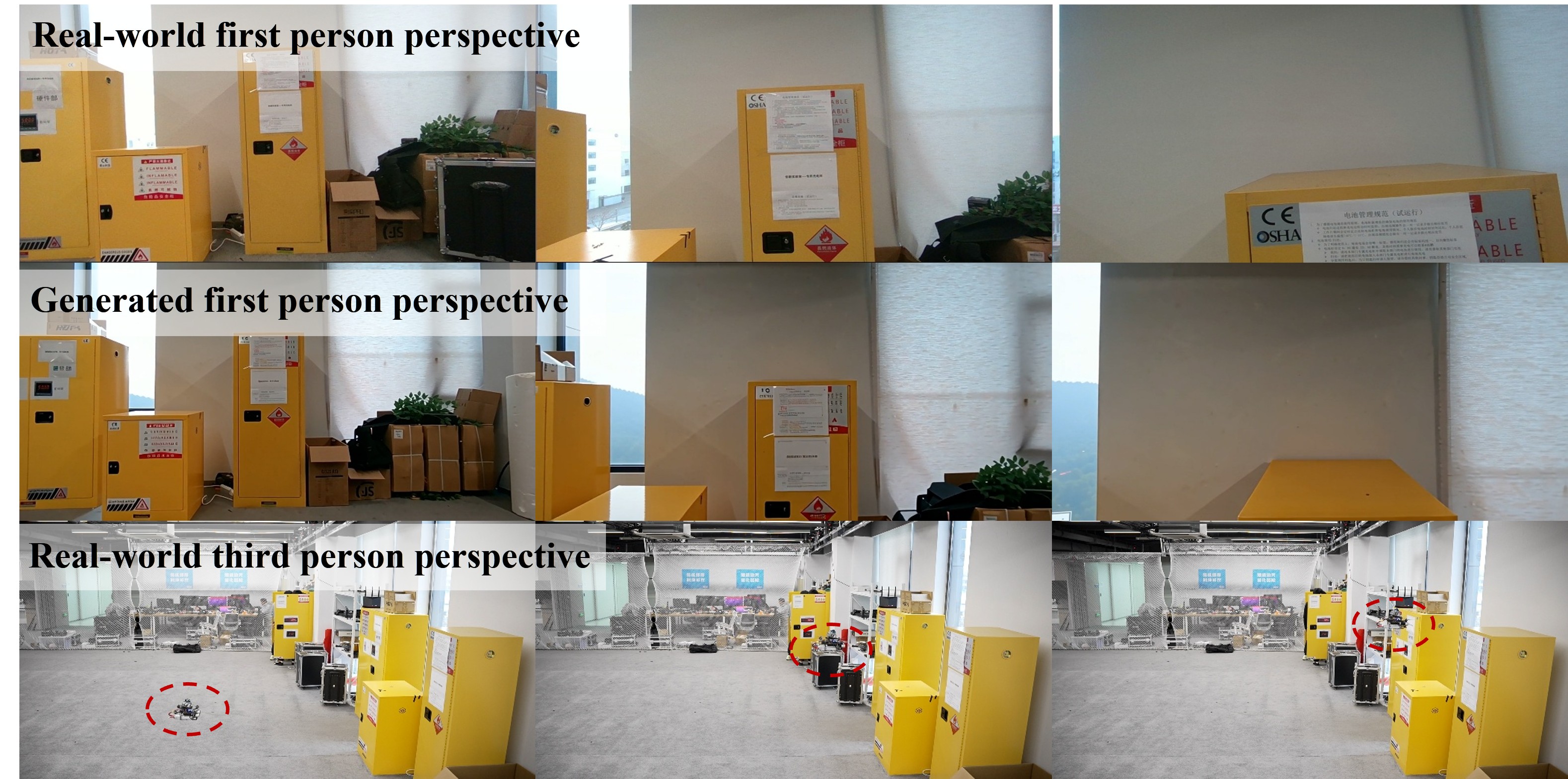}
        \caption{Move above the cabinet} 
    \end{subfigure}
    \begin{subfigure}[b]{0.9\textwidth}
        \includegraphics[width=\textwidth]{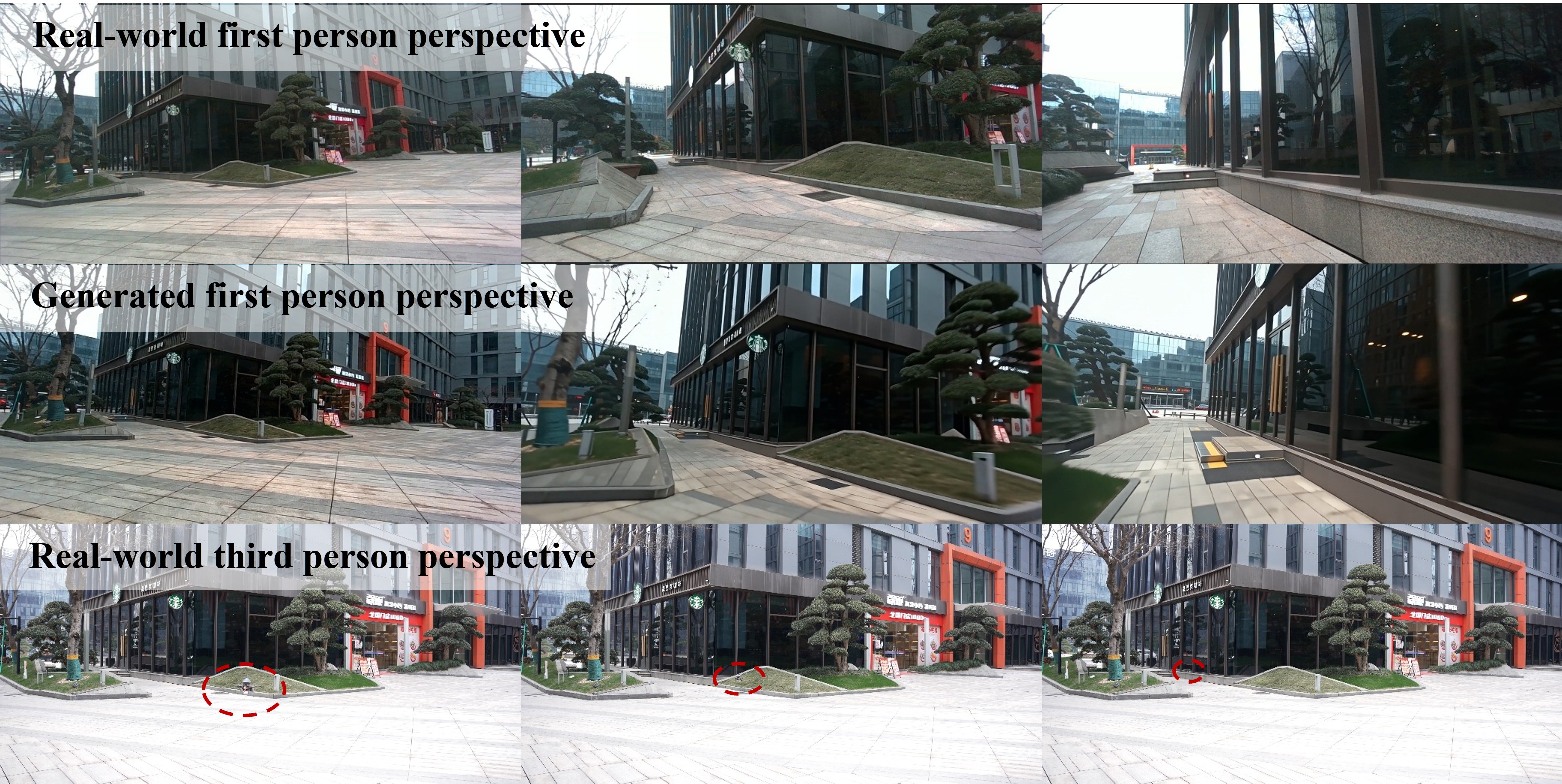}
        \caption{Navigate to the door of the starbucks} 
    \end{subfigure}
    \begin{subfigure}[b]{0.9\textwidth}
        \includegraphics[width=\textwidth]{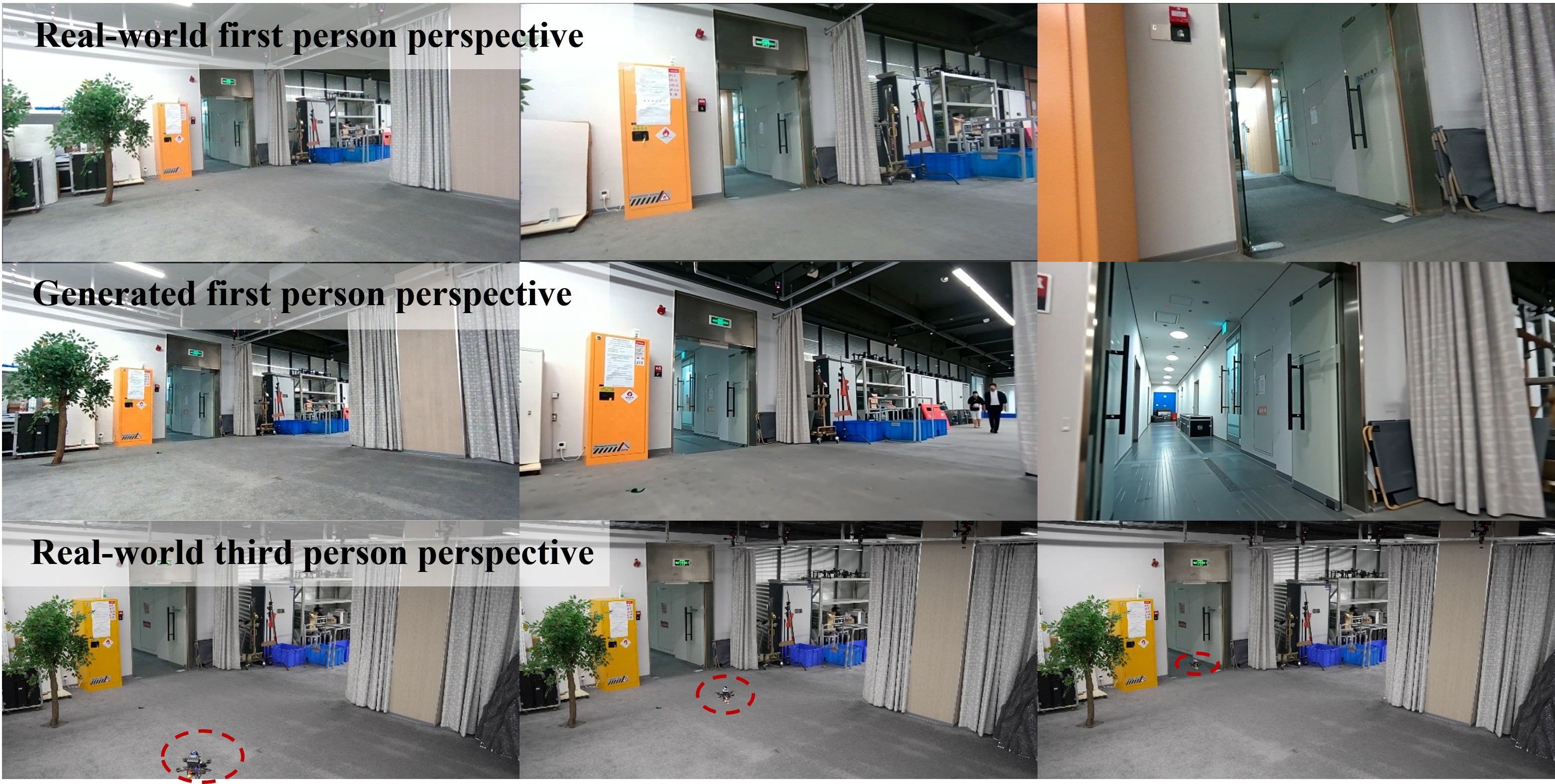}
        \caption{Find exit} 
    \end{subfigure}
    \vspace{3mm}
        
    \caption{Visual samples of our 3D navigation benchmark tasks.}
    \label{fig:task_samples}
\end{figure}

\begin{figure}[H] 
    \centering
    \begin{subfigure}[b]{0.9\textwidth}
        \includegraphics[width=\textwidth]{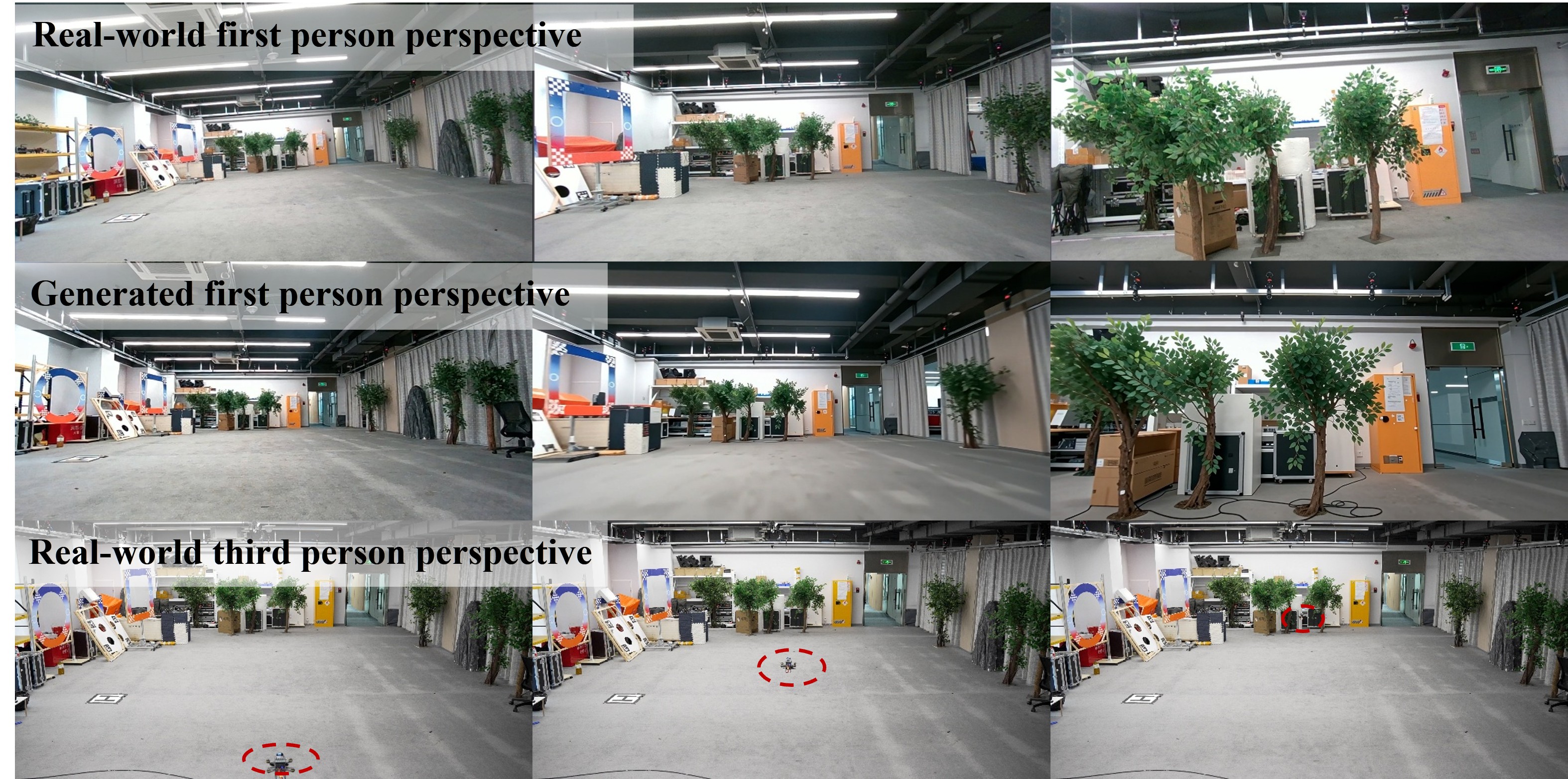}
        \caption{Fast move to the trees} 
    \end{subfigure}
    \begin{subfigure}[b]{0.9\textwidth}
        \includegraphics[width=\textwidth]{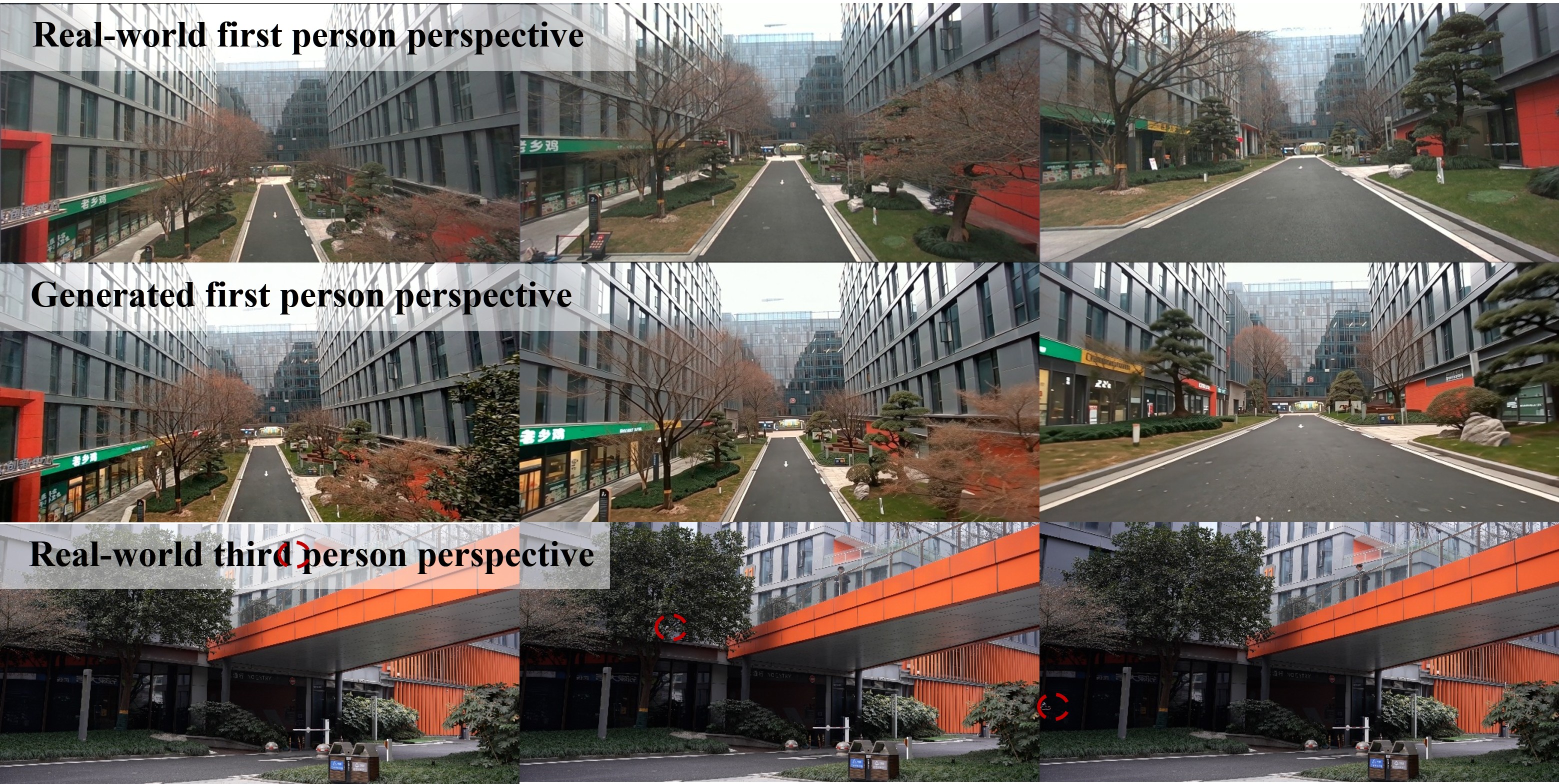}
        \caption{Fly down to the ground} 
    \end{subfigure}
    \begin{subfigure}[b]{0.9\textwidth}
        \includegraphics[width=\textwidth]{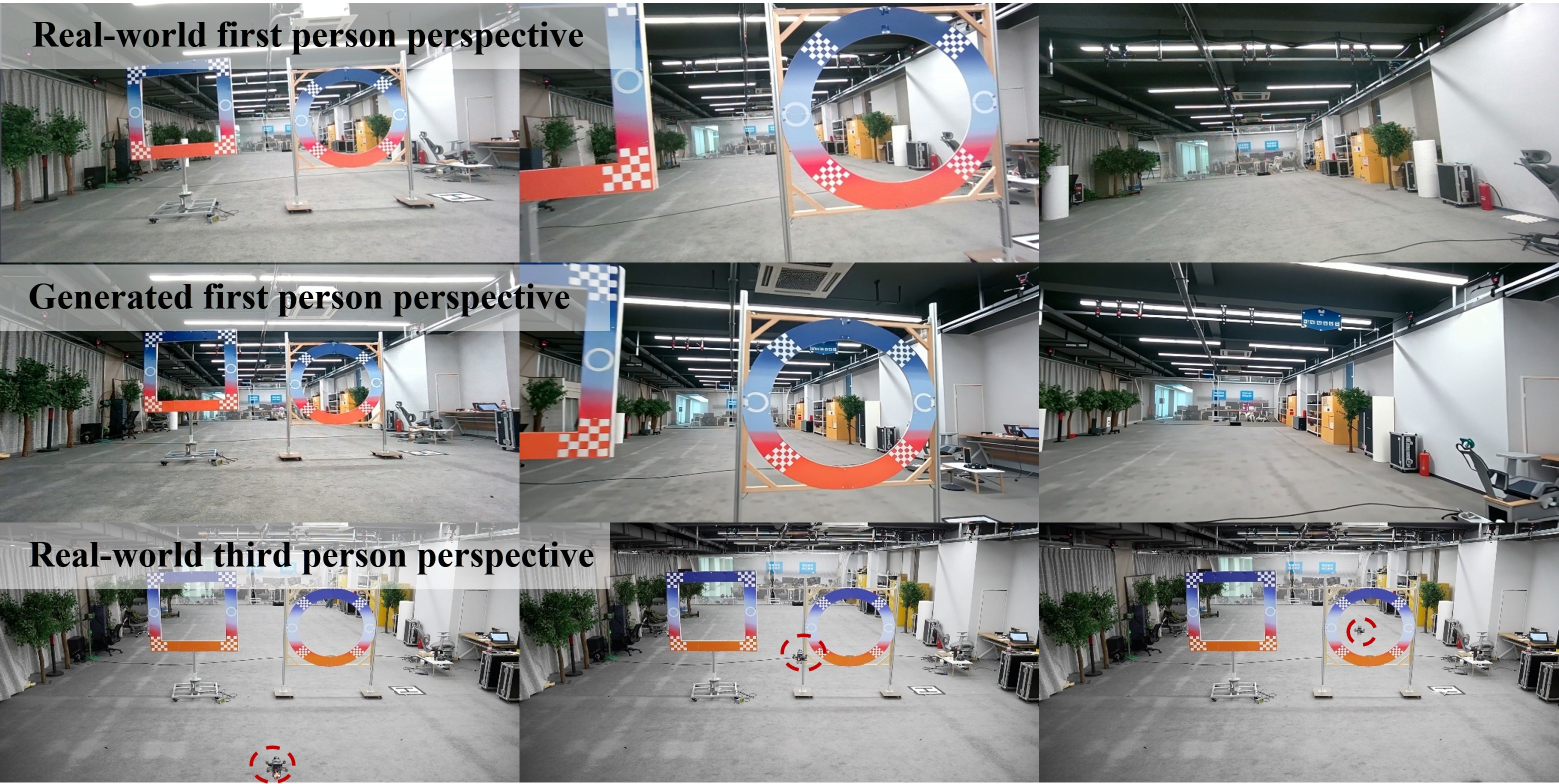}
        \caption{Fly through the middle of the right hoop.} 
    \end{subfigure}
    \vspace{3mm}
    
    \caption{Visualization of the generated videos and the real-world performance from different perspectives.}
    \label{fig:task_samples}
\end{figure}

\begin{figure}[H] 
    \centering
    \begin{subfigure}[b]{0.9\textwidth}
        \includegraphics[width=\textwidth]{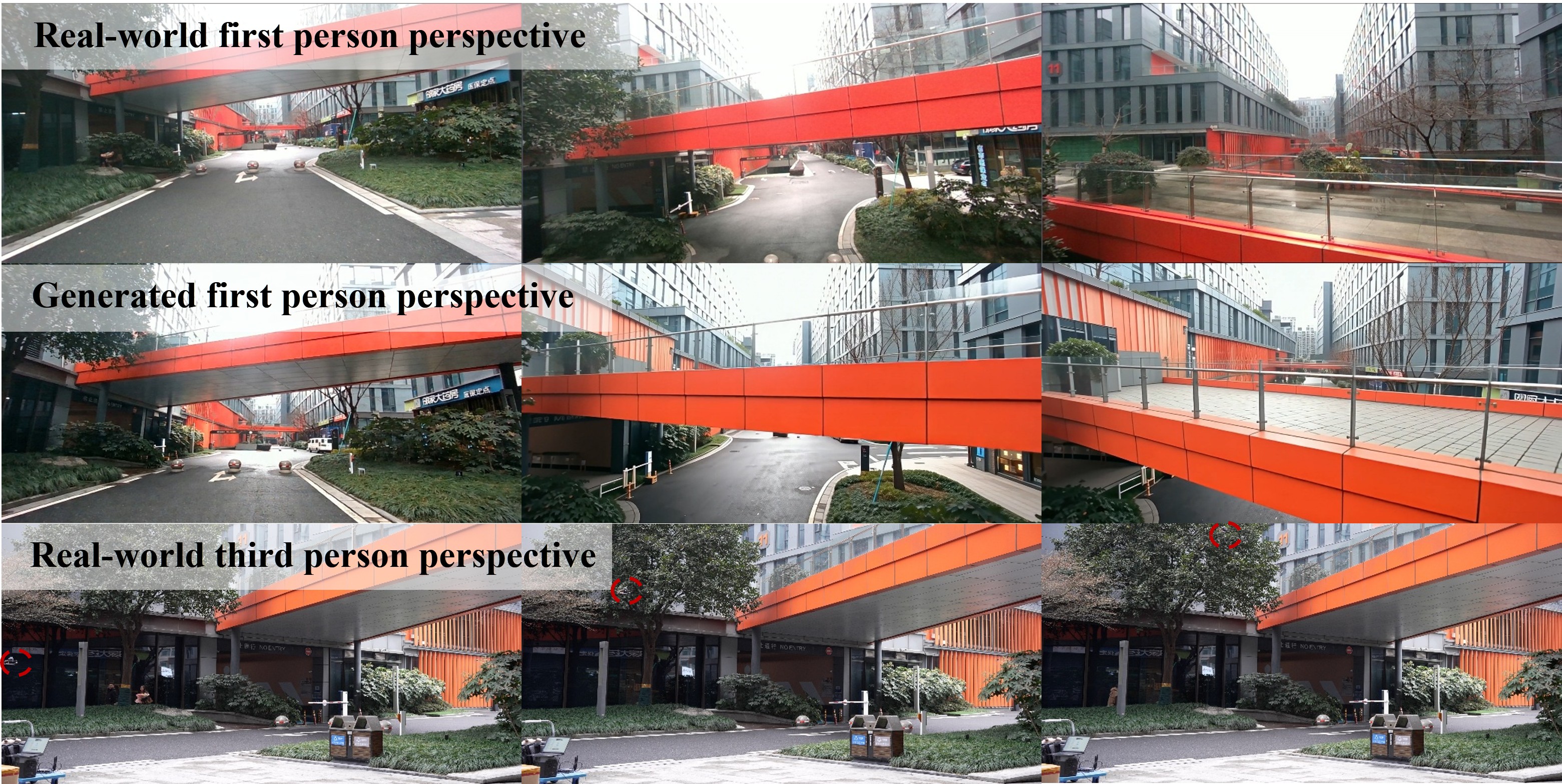}
        \caption{Fly upward and forward to the orange overpass } 
    \end{subfigure}
    \begin{subfigure}[b]{0.9\textwidth}
        \includegraphics[width=\textwidth]{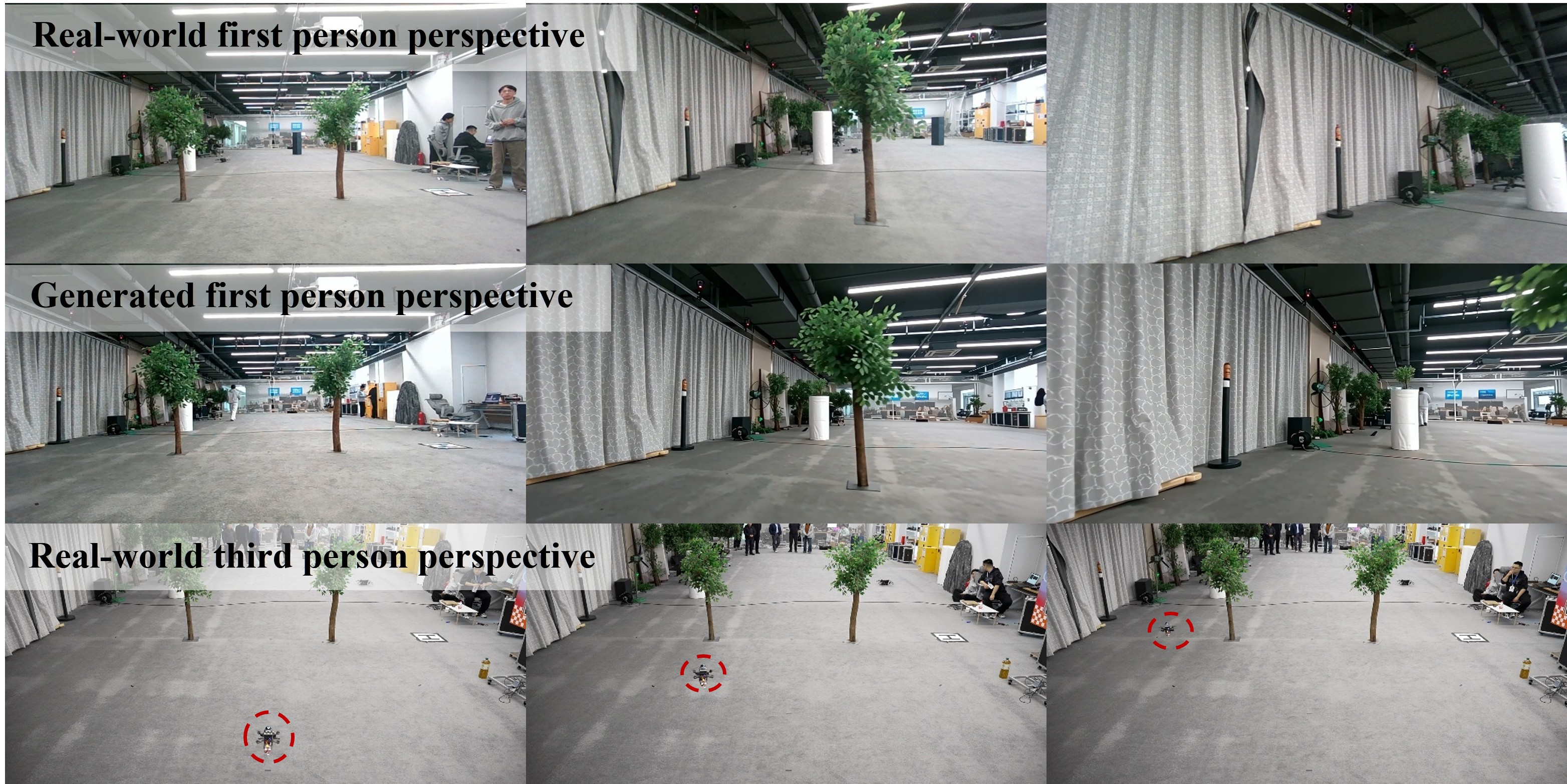}
        \caption{Move to the left of the left tree} 
    \end{subfigure}
    \begin{subfigure}[b]{0.9\textwidth}
        \includegraphics[width=\textwidth]{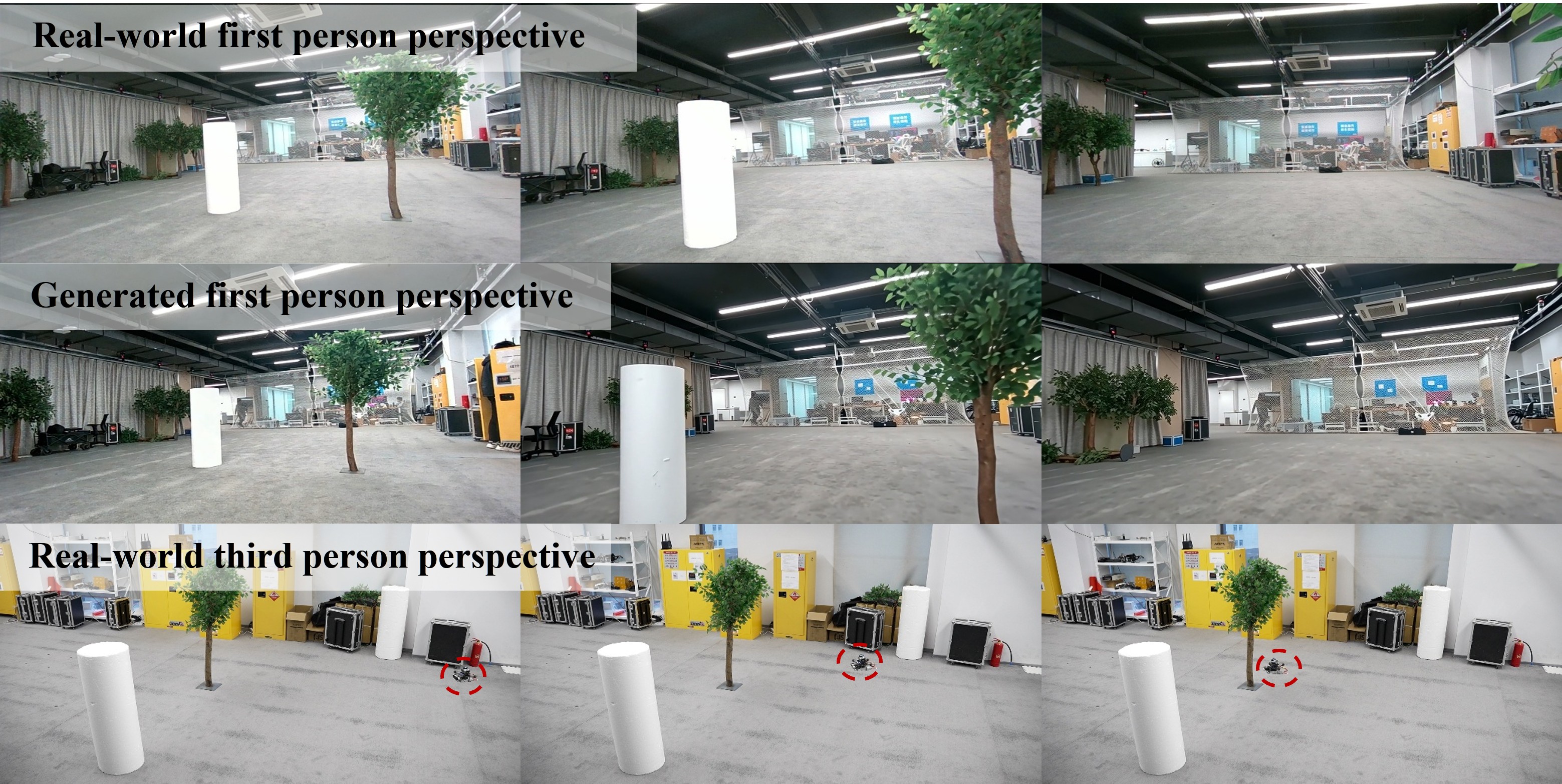}
        \caption{Move to the middle line of the column and tree} 
    \end{subfigure}
    \vspace{3mm}
    
    \caption{Visualization of the generated videos and the real-world performance from different perspectives.}
    \label{fig:task_samples}
\end{figure}

\begin{figure}[H] 
    \centering
    \begin{subfigure}[b]{0.9\textwidth}
        \includegraphics[width=\textwidth]{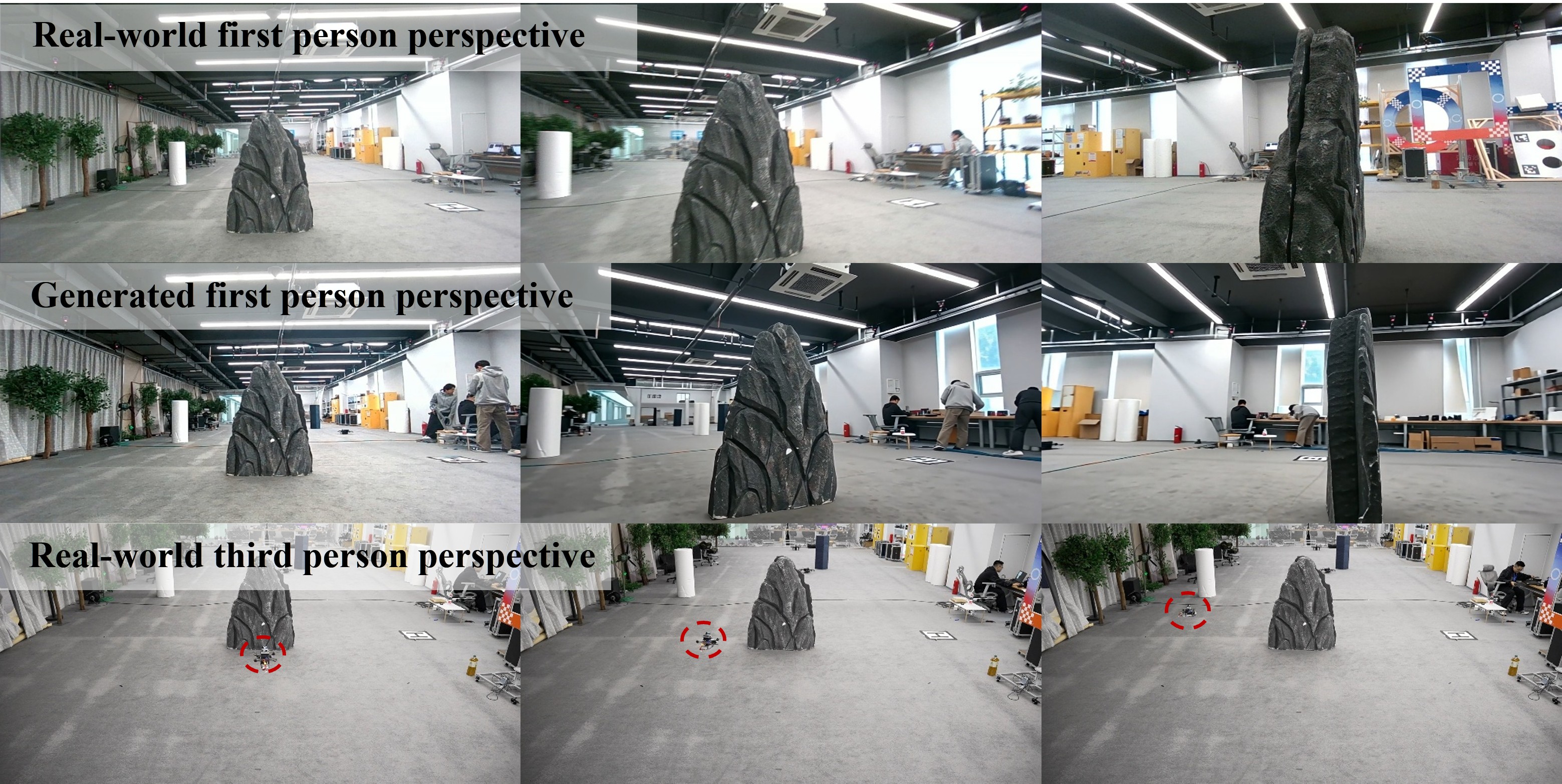}
        \caption{Move to the back of the rock} 
    \end{subfigure}
    \begin{subfigure}[b]{0.9\textwidth}
        \includegraphics[width=\textwidth]{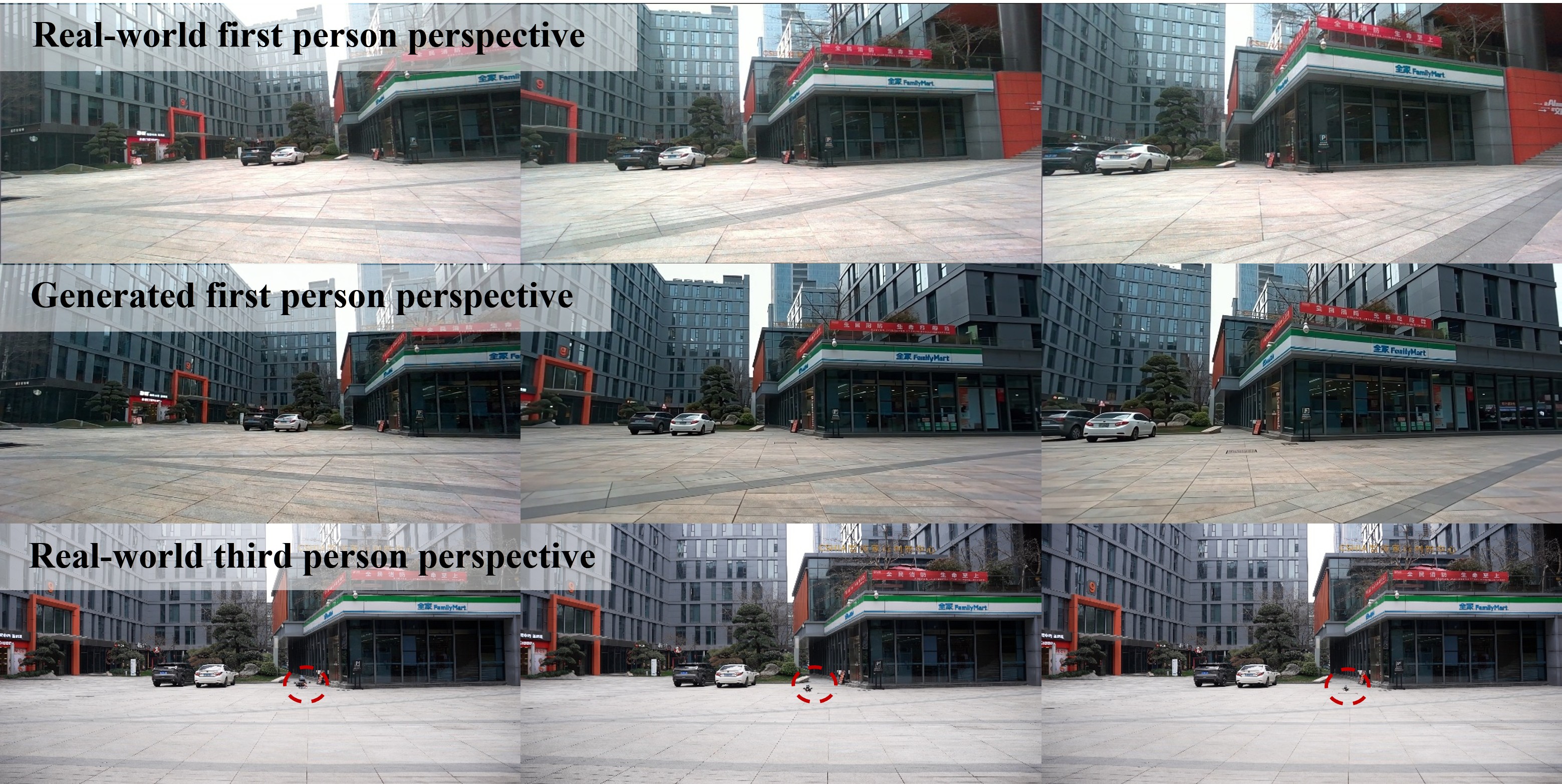}
        \caption{Move to the supermarket} 
    \end{subfigure}
    \begin{subfigure}[b]{0.9\textwidth}
        \includegraphics[width=\textwidth]{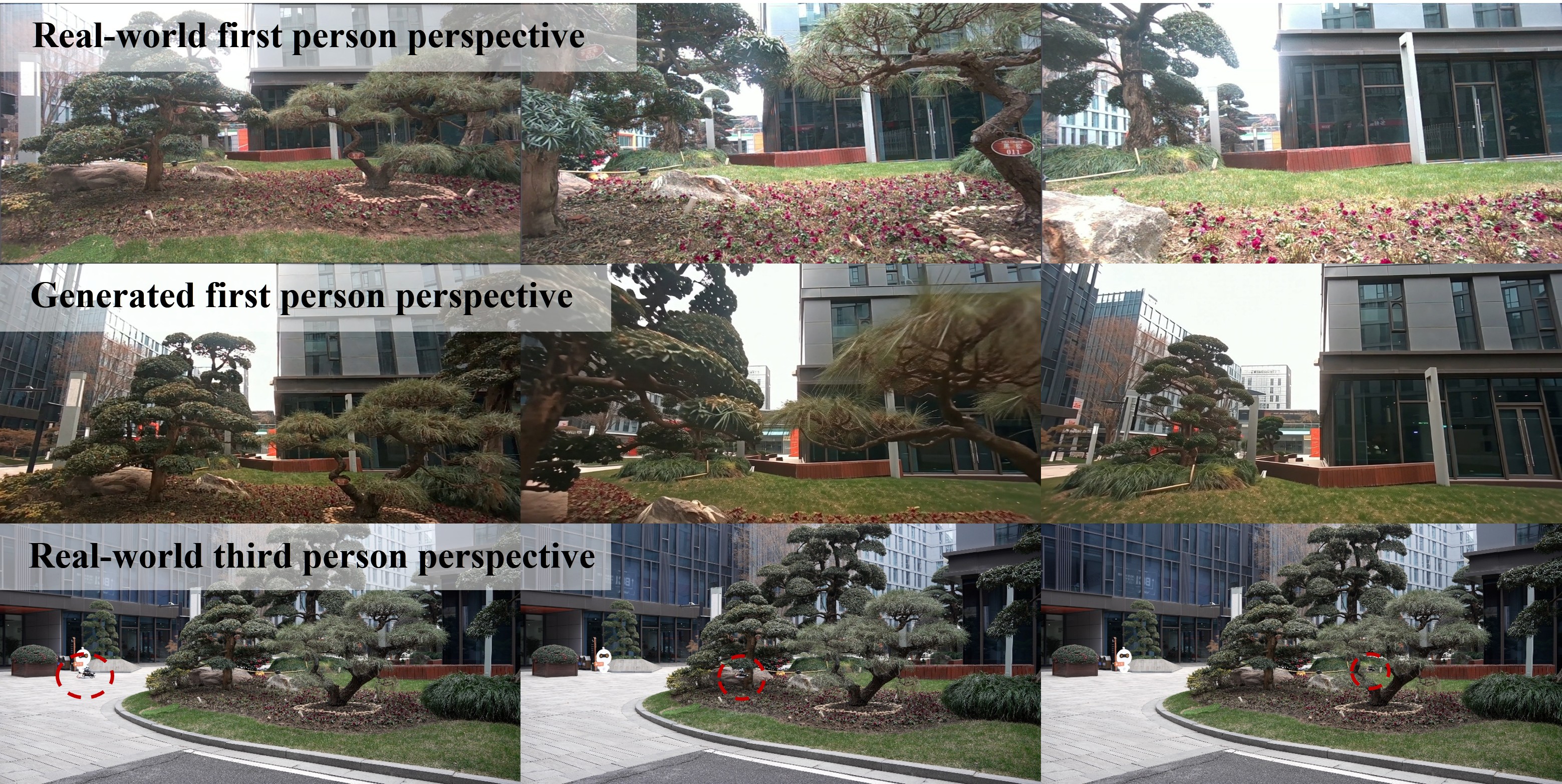}
        \caption{Fly through the gap between the two trees} 
    \end{subfigure}
    \vspace{3mm}
    
    \caption{Visualization of the generated videos and the real-world performance from different perspectives.}
    \label{fig:task_samples}
\end{figure}

\begin{figure}[H] 
    \centering
    \begin{subfigure}[b]{0.9\textwidth}
        \includegraphics[width=\textwidth]{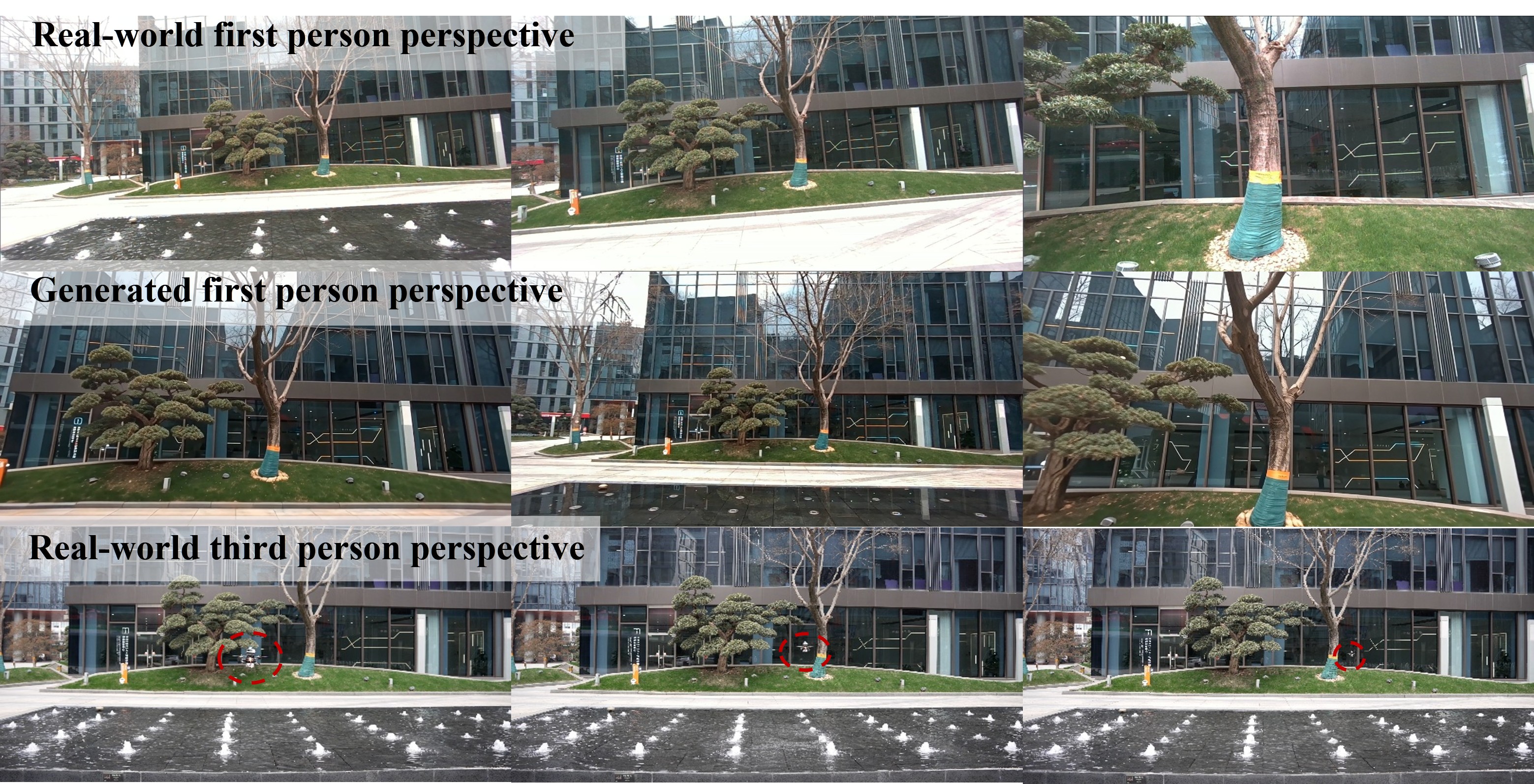}
        \caption{Move to the tree without leaves} 
    \end{subfigure}
    \vspace{3mm}

    \caption{Visualization of the generated videos and the real-world performance from different perspectives.}
    \label{fig:task_samples}
\end{figure}
\end{document}